\documentclass{article}

    \PassOptionsToPackage{numbers, compress}{natbib}

\usepackage[final]{neurips_2021}

\usepackage[utf8]{inputenc} 
\usepackage[T1]{fontenc}    
\usepackage{hyperref}       
\usepackage{url}            
\usepackage{booktabs}       
\usepackage{amsfonts}       
\usepackage{nicefrac}       
\usepackage{microtype}      
\usepackage{xcolor}         
\usepackage{amsmath}
\usepackage{graphicx}
\usepackage{algorithmic}
\usepackage{wrapfig}
\usepackage{multirow}
\usepackage{array}
\usepackage{paralist}
\usepackage{todonotes}

\usepackage{afterpage}

\usepackage[linesnumbered,ruled,vlined]{algorithm2e}
\usepackage{theoremref}

\usepackage{amssymb}
\usepackage{pifont}
\newcommand{\cmark}{\ding{51}}%

\usepackage{floatrow}
\usepackage{pgfplots}

\newcommand\Tstrut{\rule{0pt}{2.6ex}}         
\newcommand\Bstrut{\rule[-0.9ex]{0pt}{0pt}}   

\def\al{\alpha} 
\def\be{\beta}

\def\R{{\mathbb R}}

\def\exp{\operatorname{exp}}

\newcommand{\argmin}{\mathop{\mathrm{arg\,min}}}

\usepackage{soul}
\definecolor{olive}{rgb}{0.42, 0.56, 0.14}

\SetKwInput{KwInput}{Input}                
\SetKwInput{KwOutput}{Output}       
\SetKwInput{KwInit}{Init}

\newcommand{\cenmat}[1]{\hat{#1}}

\title{Accurate Point Cloud Registration \\ with Robust Optimal Transport}

\author{
  Zhengyang Shen$^*$ \\
  UNC Chapel Hill\\
  {\tt\small zyshen@cs.unc.edu}  
  \And 
  Jean Feydy\thanks{Equal Contribution.} \\
  Imperial College London \\
  {\tt\small jfeydy@ic.ac.uk}
  \And 
  Peirong Liu \\
  UNC Chapel Hill \\
  {\tt\small peirong@cs.unc.edu}
  \And 
  Ariel Hernán Curiale \\
  Harvard Medical School \\
  {\tt\small acuriale@bwh.harvard.edu}
  \And 
  Ruben San José Estépar \\
  Harvard Medical School \\
  {\tt\small rubensanjose@bwh.harvard.edu}
  \And 
  Raúl San José Estépar\\
  Harvard Medical School \\
  {\tt\small rjosest@bwh.harvard.edu}
  \And 
  Marc Niethammer \\
  UNC Chapel Hill \\
  {\tt\small mn@cs.unc.edu}
}

\begin{document}

\maketitle

\begin{abstract}
This work investigates the use of robust optimal transport (OT) for shape matching. Specifically, we show that recent OT solvers
improve both optimization-based and deep learning methods for point cloud registration,
boosting accuracy at an affordable computational cost.
This manuscript starts with a practical overview of modern OT theory. We then provide solutions to the main difficulties in using this framework for shape matching. Finally, we showcase the performance of transport-enhanced
registration models on a wide range of challenging tasks: rigid registration for partial shapes; scene flow estimation on the \text{Kitti} dataset; and nonparametric registration of lung vascular trees between inspiration and expiration. Our OT-based methods achieve state-of-the-art results on \text{Kitti} and for the challenging lung registration task, both in terms of accuracy and scalability. \\
We also release PVT1010, a new public dataset of 1,010 pairs of lung vascular trees with densely sampled points. This dataset provides a challenging use case for point cloud registration algorithms with highly complex shapes and deformations.
Our work demonstrates that robust OT enables fast pre-alignment and fine-tuning for a wide range of registration models, thereby providing a new key method for the computer vision toolbox.
Our code and dataset are available online at: \url{https://github.com/uncbiag/robot}.
\end{abstract}

\section{Introduction}
\label{sec:introduction}
\begin{figure}[h!p]
\centering
\includegraphics[width=\textwidth]{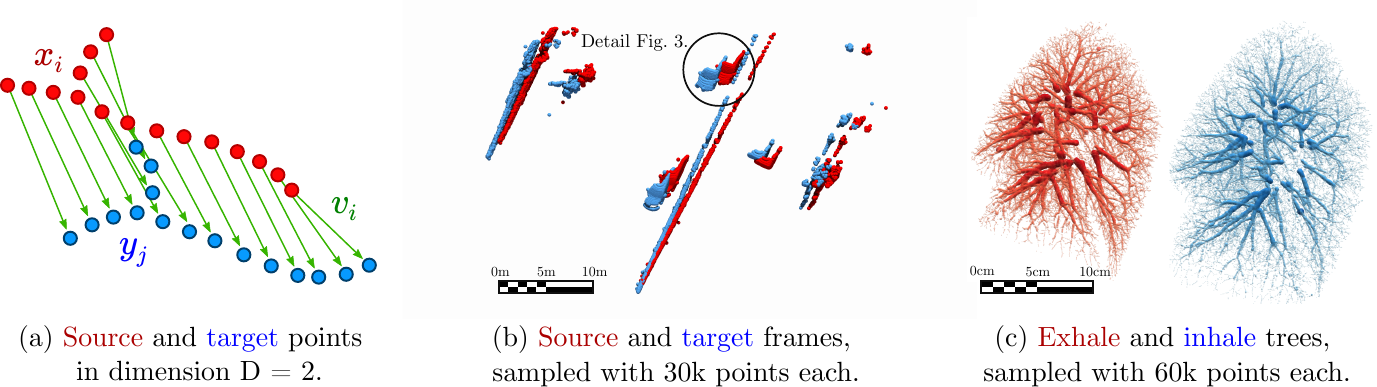}
\caption{\textbf{Robust Optimal Transport (RobOT)} generalizes sorting
to spaces of dimension $\text{D} \geqslant 1$. 
(a) RobOT is equivalent to a nearest neighbor projection
subject to mass distribution constraints that make it robust
to translations and small deformation.
We demonstrate that RobOT is now ready to be part
of the standard toolbox in computer vision
with extensive numerical experiments
for 3D scene flow estimation (b) and lung registration (c).
Rendering done with 
Paraview \cite{paraview} and PyVista \cite{pyvista}.}
\label{fig:intro}
\end{figure}

Shape registration is a fundamental but difficult problem in computer vision.
The task is to determine plausible spatial correspondences between pairs of shapes, with use cases that range from pose estimation for noisy point clouds~\cite{bouaziz2013sparse} to the nonparametric registration of high-resolution medical images~\cite{brunn2021fast}. As illustrated in Fig.~\ref{fig:intro}, most existing approaches to this problem consist of a combination of three steps, possibly fused together by some deep learning (DL) methods: (1) feature extraction; (2) feature matching; and (3) regularization using a class of acceptable transformations that is specified through a parametric or nonparametric model. 
This work discusses how tools derived from optimal transport (OT)
theory~\cite{peyre2019computational} can improve the second step of this pipeline (feature matching) on challenging problems. To put these results in context, we first present an overview of related methods.


\textbf{1. Feature extraction.} To establish spatial correspondences, one first computes descriptive local features. When dealing with (possibly annotated) point clouds,
a simple choice is to rely on Cartesian coordinates $(x, y ,z)$~\cite{arun1987least,chui2003new}. Going further, stronger descriptors capture local geometric and topological properties: examples include shape orientation and curvatures~\cite{charon2013varifold,roussillon2016kernel},
shape contexts~\cite{belongie2000shape}, spectral eigenvalues~\cite{ovsjanikov2012functional,lombaert2014spectral}
and annotations such as color~\cite{men2011color} or chemical fingerprints~\cite{gainza2020deciphering,sverrisson2021fast}.
Recently, expressive feature representations have also been \emph{learned} using deep neural networks (DNN): see~\cite{bronstein2017geometric} and subsequent works on \emph{geometric} deep learning. 
Generally, feature extractors are designed to make shape registration as unambiguous as possible. In order to get closer to the ideal case of landmark matching~\cite{bookstein1997morphometric}, we associate discriminative features to the salient points of our shapes: this increases the robustness of the subsequent \emph{matching} and \emph{regularization} steps.

\textbf{2. Feature matching.} Once computed on both of the source and target shapes, feature vectors are put in correspondence with each other.
This assignment is often encoded as an explicit mapping between the two shapes; alternatively, the vector field relating the shapes can be defined implicitly as the gradient of a geometric loss function
that quantifies discrepancies between two distributions
of features \cite{feydy2020geometric}:

\begin{enumerate}

\item[a)] A first major approach is to rely on \textbf{nearest neighbor} projections
and the related chamfer \cite{borgefors1984distance} and Hausdorff distances \cite{bouaziz2016modern},
as in the Iterative Closest Point (ICP) algorithm \cite{besl1992method}.
This method can be softened through the use
of a softmax (log-sum-exp) operator as in the many
variants of the Coherent Point Drift (CPD) method~\cite{myronenko2010point,ma2015non_tip,ma2015non_pr,gao2019filterreg},
or made robust to outliers in the specific context of rigid and affine registrations
\cite{fischler1981random,yang2020teaser,yang2019polynomial,bouaziz2013sparse}.

\item[b)] Alternatively, a second approach is to rely on \textbf{convolutional kernel norms} such as the Energy Distance \cite{rizzo2016energy},
which are also known as Maximum Mean Discrepancies (MMD) in statistics \cite{gretton2012kernel}.
These loss functions are common in imaging science \cite{pluim2001mutual}
and computational anatomy \cite{vaillant2005surface,charon2013varifold}
but are prone to vanishing gradients \cite{feydy2018global,feydy2019interpolating}.

\item[c)] Finally, a third type of approach is to rely on \textbf{optimal transport
(OT)} theory \cite{peyre2019computational}
and solutions of the earth mover's problem \cite{rubner2000earth}.
This method is equivalent to a nearest neighbor projection
under a global constraint of bijectivity
that enforces consistency in the matching.
On the one hand, OT has been known to provide
reliable correspondences in computer vision for more
than two decades \cite{chui2003new,gold1998new,kosowsky1994invisible}.
On the other hand, it has often faced major issues of scalability and robustness to outliers
on noisy data. As detailed below, the main purpose of this work is
to overcome these limitations and enable the widespread use of
OT tools for challenging registration problems.

\end{enumerate}

\textbf{3. Regularization with a deformation model.} 
The output of the two steps above
is a non-smooth vector field that may not be suitable for downstream tasks
due to e.g. tears and compression artifacts.
As a third step, most registration 
methods thus rely on regularization to obtain 
plausible deformations. This process is task-specific, with applications that range from rigid registration~\cite{zeng20173dmatch,gojcic2019perfect,deng2018ppfnet,aoki2019pointnetlk,wang2019deep,wang2019prnet} 
to free-form motion estimation~\cite{puy2020flot,wu2019pointpwc,liu2019flownet3d,gu2019hplflownet}.
In Sec.~\ref{sec:reg_model}, we address the interaction of OT
matching layers with a varied collection of regularization strategies -- from optimization-based spline and diffeomorphic models to DNNs.

\textbf{Recent progresses.}
Research works on shape registration combine ideas
from the three paragraphs above to best fit the characteristics
of computer vision problems \cite{luthi2017gaussian,donati2020deep,shen2019networks}.
Over the past few years, significant progress has been made on all fronts. 
On the one hand, (geometric) deep learning networks
have been used to define data-driven feature maps \cite{qi2017pointnet,qi2017pointnet,wang2019dynamic}
and multiscale regularization modules \cite{wu2019pointpwc,litany2017deep,shen2019region},
sometimes fused within end-to-end architectures \cite{zeng20173dmatch,puy2020flot,yew2020rpm,deng2018ppfnet}.
On the other hand, nearest neighbor projections, kernel convolutions
and transport-based matching strategies have all
been generalized to take advantage of these modern descriptors:
they can now be used in high-dimensional feature spaces \cite{johnson2019billion,feydy2020fast}.

\textbf{Challenges.}
Nevertheless, state-of-the-art (SOTA) methods in the field
still have important limitations.
First, modern deep learning pipelines are often hard to train
to ``pixel-perfect'' accuracy on non-smooth shapes, 
with diminishing returns in terms of model size and training data \cite{aoki2019pointnetlk}.
Second, scaling up point neural networks
to finely sampled shapes ($\text{N} > 10\text{k}$ points)
remains a challenging research topic \cite{gu2019hplflownet,zhou2020fully,feydy2020fast}.
Third, the impact of the choice of a specific
feature matching method on the performance of
deep learning models 
remains only partially understood \cite{huang2021comprehensive}.
\newpage

\textbf{Related works.}
Following major progress on computational OT in
the mathematical literature \cite{merigot2011multiscale,cuturi2013sinkhorn,levy2015numerical,schmitzer2019stabilized}, improved modules for feature matching have attracted interest
as a possible solution to these challenges.
Works on sliced partial OT \cite{bonneel2019spot}
and dustbin-OT \cite{dang2020learning}
have shown that outliers can be handled
effectively by OT methods for rigid registration,
beyond the classic Robust Point Matching method (RPM)~\cite{gold1998new,chui2003new}.
Going further, the Sinkhorn algorithm for entropy-regularized OT
\cite{cuturi2013sinkhorn,kosowsky1994invisible,schrodinger1932theorie,leonard2012schrodinger} has been studied extensively
for shape registration in computational anatomy \cite{feydy2017optimal,gerber2018exploratory}
and computer graphics \cite{mandad2017variance,eisenberger2020deep,pai2021fast}.
The Gromov--Wasserstein distance 
has also been used for shape analysis \cite{solomon2016entropic,vayer2019sliced},
albeit at a higher computational cost. These applications have driven interest
in the development of a complete theory for
\textbf{Robust Optimal Transport} (\text{RobOT}),
outlined in Sec.~\ref{sec:barycenter},
which handles both sampling artifacts and outliers
\cite{chizat2018unbalanced,sejourne2019sinkhorn,liero2018optimal,chizat2018scaling,sejourne2020unbalanced,mukherjee2021outlier}.
Most recently, this framework has started to be used in shape analysis with applications to shape matching \cite{feydy2017optimal},
the segmentation of brain tractograms \cite{feydy2019fast}
and deep deformation estimation with the \text{FLOT} architecture~\cite{puy2020flot}.

\textbf{Contributions.} We build upon the work above to tackle challenging point cloud registration problems for scene flow estimation and computational anatomy. Our \textbf{key contributions} are:
\begin{enumerate}
    \item \textbf{Accurate feature matching with scalable OT solvers.}
    For the first time, we scale up \text{RobOT} for deep feature matching
    to high-resolution shapes with more than 10k points.
    To this end, we leverage the latest generation of OT solvers \cite{feydy2019interpolating,feydy2020geometric}
    and overcome significant issues of memory usage and
    numerical stability. This allows us to handle fine-grained details effectively, which is key for e.g. most medical applications.
    
    \item \textbf{Interaction with task-specific regularization strategies.}
    We show how to interface \text{RobOT} matchings with advanced deformation models.
    This is in contrast with e.g. the \text{FLOT} architecture, which focuses on the direct prediction of a vector field and
    cannot be used for applications that require guarantees on the smoothness of the registration.
    
    \item \textbf{Challenging new dataset.} We release a large dataset of lung vascular trees that should be registered between inhalation and exhalation.
    This relevant medical problem involves large and complex deformations
    of high-resolution 3D point clouds.
    As a new benchmark for the community, we provide two strong
    baselines that rely respectively on global feature matching and on deep deformation estimation.
    
    \item \textbf{Consistent SOTA performance.} Our proposed models achieve SOTA results for scene flow on \text{Kitti}~\cite{menze2015joint,menze2015object} and for point-cloud-based lung registration on \text{DirLab}-\text{COPDGene}~\cite{castillo2013reference}.
    Notably, we show that \text{RobOT} is highly suited
    to fine-tuning tasks: it consistently
    turns ``good'' matchings into nearly perfect registrations
    at an affordable numerical cost.
\end{enumerate}

\textbf{Main experimental observations.}  
Is OT relevant in the deep learning era?
To answer this question decisively,
we perform extensive numerical experiments
and ablation studies.
We fully document ``underwhelming'' results
in the Supplementary Material and distill
the key lessons that we learned in the Deep-RobOT architecture
(Section~\ref{subsec:deep_deformation}).
This model relies on fast RobOT layers to cover
for the main weaknesses of point neural networks
for shape registration.
It is remarkably easy to deploy and generalizes
well from synthetic training data to real test samples.
We thus believe that it will have a stimulating impact 
on both of the computer vision and medical imaging literature.
\newpage 

\section{Robust optimal transport}
\label{sec:barycenter}

This section introduces the mathematical foundations of our work.
After a brief overview of Robust Optimal Transport (\text{RobOT})
theory, we discuss the main challenges that one encounters when using this framework for computer vision.
To avoid memory overflows and numerical stability issues,
we introduce the weighted ``\text{RobOT} matching'':
a vector field that summarizes the information
of a full transport plan with a linear memory footprint.
As detailed in the next sections, this representation lets us scale up to high-resolution shapes without compromising on accuracy.

\subsection{Mathematical background}
\label{sec:mathematical_background}

\paragraph{The assignment problem.}
If $\text{A} = (x_1, \dots, x_\text{N})$
and $\text{B} = (y_1, \dots, y_{\text{M}})$
are two \textbf{point clouds} in $\mathbb{R}^\text{3}$
with $\text{N} = \text{M}$,
the assignment problem between A and B reads:
\begin{align}
    \text{Assignment}(\text{A}, \text{B})
    ~&=~
    \min_{s : [\![1, \text{N}]\!] \rightarrow [\![1, \text{N}]\!] }
    ~~\frac{1}{2\text{N}}
    \sum_{i=1}^\text{N}
    \|x_i - y_{s(i)}\|^2_{\mathbb{R}^3}~,
    &
    \text{where $s$ is a permutation}.
    \label{eq:assignment}
\end{align}
This problem generalizes sorting to $\mathbb{R}^3$: if the points $x_i$ and $y_j$ all belong to a line, the optimal permutation
$s^*$ corresponds to a non-decreasing re-ordering of the point sets A and B \cite{peyre2019computational}.

\paragraph{Robust optimal transport.}
Further, OT theory allows us to consider problems where $\text{N} \neq \text{M}$. \textbf{Non-negative weights} $\alpha_1, \dots, \alpha_\text{N}$, $\beta_1, \dots, \beta_\text{M} \geqslant 0$ are attached to the points $x_i$, $y_j$ and account for variations of the sampling densities, while \textbf{feature vectors}
$p_1, \dots, p_\text{N}$ and $q_1, \dots, q_\text{M}$ in $\mathbb{R}^\text{D}$ may advantageously replace raw point coordinates $x_i$ and $y_j$ in $\mathbb{R}^3$.
Following \cite{liero2018optimal,chizat2018unbalanced}, the robust OT problem between the shapes $\text{A} = (\alpha_i, x_i, p_i)$ and $\text{B} = (\beta_j, y_j, q_j)$  reads:
\begin{align}\label{eq:unbalanced_Sinkhorn}
\mathrm{OT}_{\sigma, \tau}(\text{A}, \text{B}) =& \min _{(\pi_{i,j}) \in \mathbb{R}_{\geqslant 0}^{\text{N} \times \text{M}}} ~~\sum_{i=1}^\text{N} \sum_{j=1}^\text{M} \pi_{i, j} \cdot \tfrac{1}{2}\|p_i - q_j\|^2_{\R^\text{D}} \\
+~\underbrace{\sigma^2 \, \mathrm{KL}\big(\pi_{i, j} \,||\, \alpha_{i} \otimes \beta_{j}\big)}_{\text{Entropic blur at scale $\sigma$.}}  & ~+~ 
\underbrace{\tau^2 \,\mathrm{KL}\big(\textstyle \sum_j \pi_{i,j} \,||\, \alpha_{i}\big)}_{\text{$\pi$ should match A\dots }}~+~
\underbrace{\tau^2\, \mathrm{KL}\big( \textstyle \sum_i \pi_{i,j}  \,||\, \beta_{j}\big)}_{\text{\dots onto B.}}~, \nonumber
\end{align}
for any choice of the regularization parameters $\sigma > 0$ and $\tau > 0$.
In the equation above, the Kullback-Leibler divergence $\mathrm{KL}\left(a_{i} || b_{i}\right)=\sum a_{i} \log \left(a_{i} / b_{i} \right)-a_{i}+b_{i} $ is a relative entropy that penalizes
deviations of a non-negative vector of weights
$(a_i)$ to a reference measure $(b_i)$. 

\paragraph{Parameters.}
The first regularization term is scaled by the square of a \textbf{blur} radius $\sigma$. This characteristic length quantifies the fuzziness of the probabilistic assignment $(\pi_{i,j})$
between points $x_i$ and $y_j$ \cite{feydy2020geometric}.
The last two regularization terms promote the matching of the full
distribution of points A onto the target shape B:
they generalize the constraints of injectivity and surjectivity
of Eq.~\eqref{eq:assignment} to the probabilistic setting.
They are scaled by the square
of a maximum \textbf{reach} distance $\tau$:
this parameter acts as a soft upper bound on the distance
between feature vectors $p_i$ and $q_j$ that should be matched with each other \cite{feydy2019fast,sejourne2019sinkhorn}.

For shape registration, 
we use simple heuristics for the values of these
two characteristic scales:
the \textbf{blur} $\sigma$ should be equal
to the average sampling distance in feature space $\mathbb{R}^\text{D}$
while the \textbf{reach} $\tau$ should be equal
to the largest plausible displacement for any given feature vector $p_i$.
These rules are easy to follow if point features
correspond to Cartesian coordinates $x_i$ and $y_j$
in $\mathbb{R}^3$ but may lead
to unexpected behaviors if features
are output by a DNN.
In the latter case, we thus normalize
our feature vectors so that
$\|p_i\|_{\mathbb{R}^\text{D}} = \|q_j\|_{\mathbb{R}^\text{D}} = 1$
and pick values for $\sigma$ and $\tau$ between $0$ and $2$.


\paragraph{Working with a soft, probabilistic transport plan.}
As detailed in \cite{feydy2020geometric}, 
scalable OT solvers for Eq.~\eqref{eq:unbalanced_Sinkhorn} return a pair of dual vectors
$(f_i) \in \mathbb{R}^\text{N}$ and 
$(g_j) \in \mathbb{R}^\text{M}$
that encode \textbf{implicitly} an optimal transport plan
$(\pi_{i,j}) \in \mathbb{R}^{\text{N}\times\text{M}}$
with coefficients:
\begin{align}\label{eq:trans_plan}
    \pi_{i,j}
    ~=~
    \alpha_i \beta_j\,\cdot\,
    \exp \tfrac{1}{\sigma^2}\big[ f_i + g_j - \tfrac{1}{2}\|p_i- q_j\|^2_{\R^\text{D}} \big]~\geqslant~0~.
\end{align}
In the limit case where $\sigma$ tends to $0$ and $\tau$ tends to $+\infty$, for generic point clouds $(x_i)$ and $(y_j)$
with $\text{N} = \text{M}$ and equal weights $\alpha_i = \beta_j = 1 / \text{N}$, $\pi_{i,j}$ is a permutation matrix \cite{peyre2019computational}. We retrieve the simple assignment problem of Eq.~\eqref{eq:assignment}: $\pi_{i,j} = 1/\text{N}$ if $j = s^*(i)$ and $0$ otherwise.
However, in general the transport plan must be understood as a probabilistic map between the point distributions A and B that assigns
a weight $\pi_{i,j}$ to the coupling ``$x_i \leftrightarrow y_j$''.
For shape registration, 
this implies that the main difficulties for using robust OT are two-fold:
first, the coupling $\pi$ is not one-to-one, but one-to-\emph{many};
second, the lines and columns of the transport plan $\pi$ do not sum up to one. Notably, this implies that when $\tau < +\infty$,
the gradient of the OT cost with respect to the point positions $x_i$ is not homogeneous: we observe vanishing and inflated values across the domain
\cite{sejourne2019sinkhorn}.

\subsection{RobOT: a convenient representation of the optimal transport plan}
\label{sec:robot_matching}

\paragraph{The weighted RobOT matching.}
To work around these issues, we introduce the vector field:
\begin{align} \label{eq:robot_matching}
    v_i
    ~&=~ \frac{\sum_{j=1}^\text{M} \pi_{i,j} \cdot (y_j - x_i) }{\sum_{j=1}^\text{M} \pi_{i,j}}~\in~\mathbb{R}^3
    & 
    \text{with confidence weights}
    & &
    w_i~&=~ \sum_{j=1}^\text{M} \pi_{i,j}~\geqslant~0~.
\end{align}
This object has the same memory footprint as the input shape $\text{A}$ and summarizes the information that is contained
in the N-by-M transport plan $(\pi_{i,j})$ -- a matrix that
is often too large to be stored and manipulated efficiently.
This ``weighted \text{RobOT} matching'' is at the heart
of our approach and generalizes the standard Monge map
from classical OT theory \cite{peyre2019computational}
to the setting of (deep) shape registration.
In practice, the weighted vector field $(w_1, v_1), \dots, (w_\text{N}, v_\text{N})$
is both convenient to use and easy to compute on GPUs. Let us briefly explain why.

\paragraph{Fast implementation.}
Our differentiable \text{RobOT} layer takes as input
the two shapes $\text{A} = (\alpha_i, x_i, p_i)$ and $\text{B} = (\beta_j, y_j, q_j)$,
with feature vectors $p_i$ and $q_j$ in $\mathbb{R}^\text{D}$ that have been computed upstream using e.g. a point neural network.
It returns the $\text{N}$ vectors $v_i$ with weights $w_i$
that map the source points $x_i$ onto the targets $y_j$ in $\mathbb{R}^3$.
Starting
from the input point features $p_i$, $q_j$
and weights $\al_i$, $\be_j$,
we first compute the optimal dual vectors
$f_i$ and $g_j$ using the fast solvers
of the \text{GeomLoss} library \cite{feydy2019interpolating}.
We then combine  Eq.~\eqref{eq:trans_plan}
with Eq.~\eqref{eq:robot_matching} to compute the \text{RobOT}
vectors $v_i$ and weights $w_i$
with $O(\text{N}+\text{M})$ memory footprint
using the \text{KeOps} library \cite{charlier2021kernel,feydy2020fast}
for \text{PyTorch} \cite{paszke2017automatic} and \text{NumPy} \cite{van2011numpy}.
We use a log-sum-exp formulation to ensure
numerical stability.
Remarkably,
our implementation scales up to $\text{N}, \text{M} = 100\text{k}$
in fractions of a second.
Unlike common strategies that are based on dense
or sparse representations of the transport plan $\pi$,
our approach is perfectly suited
to a \emph{symbolic} implementation \cite{feydy2020fast}
and streams well on GPUs with optimal, contiguous
memory accesses.

\paragraph{Comparison with nearest neighbor projections.}
We use our \text{RobOT} layer as a plug-in replacement
for closest point matching \cite{besl1992method}.
The \emph{blur} ($\sigma$) and 
\emph{reach} ($\tau$) scales 
play similar roles to  the standard deviation ($\sigma$)
and weight of the uniform distribution ($w$)
in the Coherent Point Drift (CPD) method \cite{myronenko2010point}:
they allow us to smooth the matching in order
to increase robustness to sampling artifacts.

The main difference
between projection-based matching and \text{RobOT}
is that the latter enforces a 
mass distribution constraint
between the source and the target.
This prevents our matching vectors from accumulating 
on the boundaries of the distributions of
point features $(p_1, \dots, p_\text{N})$ 
and $(q_1, \dots, q_\text{M})$ in $\mathbb{R}^\text{D}$ \cite{feydy2018global}.
This property is most desirable when 
the shapes to register are fully observed,
with a dense sampling: 
as detailed in Suppl.~\ref{sec:robot_vs_nn},
enforcing the \textbf{global consistency}
of a matching is then a worthwhile registration prior.

{\bf Partial registration.} 
On the other hand, we must also stress that 
OT theory has known limitations \cite{feydy2020geometric}.
First of all, the \text{RobOT} matching cannot
guarantee the preservation of remarkable points
or of the shapes' topologies ``on its own'':
it should be combined with relevant feature extractors and regularizers.
Going further,
partial registration is not a natural fit for standard OT formulations
which assume that \emph{all} points from both shapes
must be put in correspondence with each other. 

To mitigate this issue, \text{RobOT} leverages the theory
of \emph{unbalanced} optimal transport \cite{chizat2018unbalanced,sejourne2019sinkhorn,liero2018optimal,chizat2018scaling,sejourne2020unbalanced}:
we rely on \emph{soft} Kullback-Leibler penalties
to enforce a matching between the shapes $\text{A}$ and $\text{B}$
in Eq.~(\ref{eq:unbalanced_Sinkhorn}).
In practice, the \text{RobOT} confidence weights $w_i$
of Eq.~(\ref{eq:robot_matching})
act as an attention mechanism:
they vanish when no target feature vector $q_j$
can be found in a $\tau$-neighborhood 
of the source vector $p_i$ in $\mathbb{R}^\text{D}$,
where $\tau$ is the \emph{reach} scale
that is associated to Eq.~(\ref{eq:unbalanced_Sinkhorn}).
This lets our registration method focus 
on reliable matches between similar features,
without being fooled
by strong constraints of bijectivity.
As detailed in Suppl.~\ref{sec:suppl_fea_matching},
combining standard \text{FPFH} features~\cite{rusu2009fast}
with the rigid
projection of Eq.~(\ref{eq:rigid})
allows us to register partially observed shapes
that have little overlap with each other.

\section{Regularization and integration with a deep learning model}
\label{sec:reg_model}

\subsection{Smooth-RobOT: algebraic and optimization-based regularization}

\label{subsec:reg_fea_match}

\label{subsec:lung_fea_learning}

\textbf{Notations.} 
We now detail how to interface the
weighted \text{RobOT} matching with regularization models 
and feature extractors
that may be handcrafted~\cite{tombari2010unique,rusu2008aligning,rusu2009fast} or learnt using a deep neural network~\cite{zeng20173dmatch,gojcic2019perfect,deng2018ppfnet,deng2018ppf}.
Recall that we intend to register a source point cloud $x_1, \dots, x_\text{N}$ onto a target $y_1, \dots, y_\text{M}$ in $\mathbb{R}^3$.
Non-negative weights $\alpha_1, \dots, \alpha_\text{N}$
and $\beta_1, \dots \beta_\text{M} \geqslant 0$ let
us take into account variations in the sampling densities
and we assume that
point features $p_1, \dots, p_\text{N}$,
$q_1, \dots, q_\text{M}$ in $\mathbb{R}^\text{D}$
have been computed upstream by a relevant
feature extractor.
For every source point $x_i$, the \text{RobOT}
matching layer then provides
a \textbf{desired displacement} $v_i$ in $\mathbb{R}^3$
with \textbf{influence weight} $w_i \geqslant 0$.

\textbf{Smoothing in closed form.} 
Standard computations let us
derive closed-form expressions for rigid and affine registration \cite{horn1987closed,myronenko2010point,arun1987least}.
These respectively correspond to transformations:
\begin{alignat}{8}
    \text{(Rigid RobOT)}\qquad&& x\in \mathbb{R}^{3} ~~&\mapsto~~ &(x - x_c)\, & U V^\top &\,+\, x_c + v_c \in \mathbb{R}^{3}~, \label{eq:rigid}\\
    \text{(Affine RobOT)}\qquad&& x\in \mathbb{R}^{3} ~~&\mapsto~~ & (x - x_c)\, &(\cenmat{X}^\top W \cenmat{X})^{-1}(\cenmat{X}^\top W \cenmat{Y}) &\,+\, x_c + v_c \in \mathbb{R}^{3}~,\label{eq:affine}
\end{alignat}
where $W = \text{Diag}(w_i) \in \mathbb{R}^{\text{N}\times\text{N}}$ is the diagonal matrix of influence weights,
$x_c = \sum_i w_i x_i / \sum_i w_i \in \mathbb{R}^3$ is the barycenter
of the source shape,
$v_c = \sum_i w_i v_i / \sum_i w_i \in \mathbb{R}^3$ is the
average desired displacement,
$\cenmat{X} = (x_i - x_c) \in \mathbb{R}^{\text{N}\times 3}$ 
is the centered matrix of source positions,
$\cenmat{Y} = (x_i + v_i - x_c - v_c) \in \mathbb{R}^{\text{N}\times 3}$
is the centered matrix of desired targets and
$USV^\top$ is the singular value decomposition of  $\cenmat{X}^\top W \cenmat{Y} \in \mathbb{R}^{3\times 3}$. 
This corresponds to a weighted Kabsch algorithm \cite{kabsch1976solution}.
Likewise, we implement free-form spline registration
using the \text{KeOps} library \cite{feydy2020fast,charlier2021kernel}.
A Nadaraya--Watson interpolator
with kernel $k : (x,y) \in \mathbb{R}^3 \times \mathbb{R}^3 \mapsto k(x,y) > 0$ \cite{nadaraya1964estimating,watson1964smooth}
induces a transformation:
\begin{align}
    \text{(Spline RobOT)}\qquad x\in\mathbb{R}^3~
    \mapsto~
    x ~+~
    \textstyle\sum_{i=1}^\text{N} w_i k(x_i, x) v_i ~/~ \textstyle\sum_{i=1}^\text{N} w_i k(x_i, x) \in \mathbb{R}^3~.
    \quad~~\label{eq:spline}
\end{align}


\label{sec:global_feature_matching}

\textbf{Black-box deformation models.} Going further,
we interface
\text{RobOT} matchings with arbitrary deformation modules
$\text{Morph}:(\theta,x_i) \mapsto \hat{y}_i \in \mathbb{R}^3$
parameterized by a vector $\theta$ in $\mathbb{R}^\text{P}$.
If $\text{Reg}(\theta)$ denotes
a regularization penalty on the parameter $\theta$
(e.g. a squared Euclidean norm),
we use standard optimizers such as
L-BFGS-B \cite{lbfgs} and 
Adam \cite{adam} 
to find the optimal deformation parameter:
\begin{align}
    \theta^*=\argmin_{\theta\in\mathbb{R}^\text{P}}~~ \text{Reg}(\theta) + \textstyle\sum_{i=1}^{\text{N}} w_i \|x_i + v_i - \text{Morph}(\theta, x_i)\|_{\mathbb{R}^3}^2\,.
    \label{eq:morph_optimization}
\end{align}
This optimization-based method is especially 
relevant in the context of computational anatomy,
where smooth and invertible deformations
are commonly defined through the
Large Deformation Diffeomorphic Metric Mapping (LDDMM) framework~\cite{beg2005computing,bone2018deformetrica,feydy2020geometric}.
We stress that different transformation models may result in different registration results and refer to Suppl.~\ref{sec:suppl_fea_matching} for further details.

Optimization-based approaches provide strong geometric guarantees on the final matching.
But unfortunately, these often come 
at a high computational price:
to register complex shapes, 
quasi-Newton optimizers require dozens of evaluations of the
deformation model $\text{Morph}(\theta,x)$ and of its gradients.
In practice, 
fitting a complex model to a pair of high-resolution
shapes may thus take several minutes or seconds \cite{brunn2021fast}.
This precludes real-time processing
and hinders research on advanced deformation models.

\subsection{Deep-RobOT: registration via deep deformation prediction}
\label{sec:reg_deep_deform}
\label{subsec:deep_deformation}

\textbf{Deformation prediction.}
%
In this context, there is growing interest
in \emph{fast} learning methods 
that avoid the use of iterative optimizers.
The idea is to train a deep neural network
$\text{Pred}: (x_i, y_j) \mapsto \theta$
that takes as input two point clouds $\text{A} = (x_1, \dots, x_\text{N})$, $\text{B} = (y_1, \dots, y_\text{M})$ and \textbf{directly predicts the optimal vector of parameters $\theta$} 
for a transformation model $\text{Morph}(\theta,\,\cdot\,)$
that should map $\text{A}$ onto $\text{B}$.
Assuming that the prediction network $\text{Pred}$
has been trained properly,
this strategy enables real-time processing
while leveraging the task-specific geometric priors 
that are encoded
within the deformation model.

Over the last few years, numerous authors have
worked in this direction for rigid registration~\cite{zeng20173dmatch,gojcic2019perfect,deng2018ppfnet,aoki2019pointnetlk,wang2019deep,wang2019prnet} and scene flow estimation~\cite{puy2020flot,wu2019pointpwc,liu2019flownet3d,gu2019hplflownet}.
Comparable research on diffeomorphic models
has focused on images that are supported on
a dense grid, with successful applications to e.g. the registration of 3D brain volumes \cite{yang2017quicksilver,balakrishnan2019voxelmorph,shen2019networks}.
As of 2021, prediction-based approaches have thus become
standard methods for 3D shape registration.

\newpage

\begin{figure}[t]
\includegraphics[width=1\textwidth]{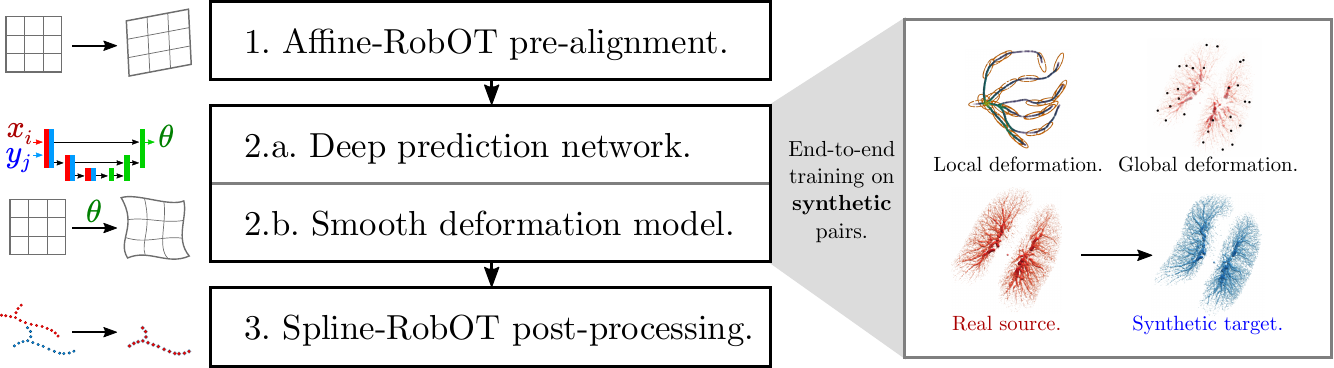}
\caption{\textbf{The D-RobOT architecture.} 
\textbf{Left:} We apply three successive registration modules that bring the
moving source shape increasingly close to the
fixed target point cloud.
On the one hand,
the RobOT-based pre-alignment (1) and fine-tuning (3) steps
take the $(x,y,z)$ coordinates as input features
and do not require any training.
On the other hand, our deep registration module (2)
relies on a multi-scale point neural network 
``$\text{Pred} : (x_i,y_j) \mapsto \theta$'' (2.a)
and a task-specific deformation model ``$\text{Morph}(\theta,x_i) \mapsto \hat{y}_i$'' (2.b).
We train it end-to-end on a dataset of synthetic pairs
of shapes with known ground truth correspondences.
\textbf{Right:} To generate these pairs,
we apply random deformations to real source shapes.
For lung registration, we apply successively
a vessel-preserving local perturbation
and a smooth global deformation -- as detailed in Suppl.~\ref{sec:suppl_synth_data}.
}
\label{fig:flow_pipeline}

\end{figure}

\textbf{The D-RobOT model.} 
In practice though, prediction methods still face three major challenges: 
\begin{compactenum}
    \item Common architectures may not be equivariant to \textbf{rigid or affine transformations}.
    \item Dense 3D annotation is expensive, especially in a medical context. 
    As a consequence, most predictors are trained on \textbf{synthetic data} and
    have to overcome a sizeable \textbf{domain gap}.
    \item Predicted registrations may be \textbf{less accurate} than optimization-based solutions. 
    Training registration networks to pixel-perfect
    accuracy is notoriously hard, with diminishing returns in terms
    of model size and number of training samples.
\end{compactenum}
We propose to address these issues using
RobOT layers for pre-alignment and post-processing.
Our Deep \text{RobOT} (\text{D-RobOT}) model is an
end-to-end deep learning architecture
that is made up of three consecutive steps 
that we illustrate in Figure~\ref{fig:flow_pipeline} and detail in Suppl.~\ref{sec:suppl_deep_deform}:

\begin{compactenum}
    \item  {\bf OT-based  pre-alignment.} 
We use the rigid or affine S-RobOT models
of Eqs.~(\ref{eq:rigid}-\ref{eq:affine}) to normalize the pose of the source shape. This is a fast
and differentiable pre-processing.
\item {\bf Deep registration module.} 
We combine a deep predictor
with a task-specific deformation model to register 
the pre-aligned source onto the target.
For the prediction network 
$\text{Pred}:(x_i,y_j) \mapsto \theta$,
we use a multiscale point neural network that is adapted
from the \text{PointPWC-Net} architecture~\cite{wu2019pointpwc}.
We refer to Suppl.~\ref{sec:suppl_deep_deform} for a full
description of our architecture and training loss.
\item {\bf OT-based post-processing.} 
In order to reach ``pixel-perfect'' accuracy,
we use the spline S-RobOT deformation model of Eq.~(\ref{eq:spline})
with a task-specific kernel $k(x,y)$.
\end{compactenum}



\textbf{Complementary strengths and weaknesses.} 
We apply these three steps successively,
which brings the moving source $\text{A} = (x_1, \dots, x_\text{N})$
increasingly close to the fixed target
$\text{B} = (y_1, \dots, y_\text{M})$.
Remarkably, each step of our method
covers for the weaknesses of the other modules:
the RobOT-based 
\textbf{pre-alignment} makes our pipeline robust
to changes of the 3D acquisition parameters;
our multiscale neural \textbf{predictor} is able
to match corresponding key points quickly,
even in complex situations;
the domain-specific \textbf{deformation model} acts as a regularizer
and improves the generalization properties of the deep registration module;
the RobOT-based \textbf{fine-tuning} improves accuracy
and helps our model to
overcome the domain gap between synthetic and real shape data.

As detailed below, the D-RobOT model generalizes well outside
of its training dataset and outperforms
state-of-the-art methods on several challenging problems.
We see it as a pragmatic architecture for shape registration,
which is easy to deploy and tailor to domain-specific
requirements.
As discussed in Suppl.~\ref{sec:suppl_deep_deform},
we found that introducing sensible
geometric priors through our RobOT layers and 
the deformation model
$\text{Morph}:(\theta,x_i) \mapsto \hat{y}_i$
results in a ``forgiving'' pipeline: 
\textbf{our model produces accurate results,
even when trained on synthetic data that is not very realistic}.
In a context where generating plausible 3D deformations
is easier than developing custom registration models
for every single task (e.g. in computational anatomy),
we believe that this is an important observation.

\section{Scene flow estimation}

 {\bf Benchmark.} 
We now evaluate our method on a standard
 registration task in computer vision: the estimation of scene flow
 between two successive views of the same 3D scene.
 We follow the same experimental setting as in~\cite{wu2019pointpwc, gu2019hplflownet}, with full details provided in Suppl.~\ref{sec:suppl_kitti}: 
 \begin{compactenum} 
 \item We train on the synthetic \textbf{Flying3D} dataset~\cite{mayer2016large},
 which is made up of multiple moving objects that are sampled at random from \text{ShapeNet}. We take 19,640 pairs of point clouds for training, with dense ground truth correspondences. 
 \item We evaluate on 142 scene pairs from \textbf{Kitti}, a real-world dataset \cite{menze2015joint,menze2015object}. We conduct experiments using 8,192 and 30k points per scan, sampled at random from the original data.
 \end{compactenum}
 
 \textbf{Performance metrics.}
 We evaluate all methods as in~\cite{wu2019pointpwc}: \textbf{EPE3D} is the average 3D error, in centimeters; \textbf{Acc3DS} is the percentage of points with 3D error\,$<$\,5\,cm or relative error\,$<$\,5\%; 
 \textbf{Acc3DR} is the percentage of points with 3D error\,$<$\,10\,cm or relative error\,$<$\,10\%;
 \textbf{Outliers3D} is the percentage of points with 3D error\,$>$\,30\,cm or relative error\,$>$\,10\%;
 \textbf{EPE2D} is the average 2D error obtained by projecting the point clouds onto the image plane, measured in pixels;
 \textbf{Acc2D} is the percentage of points with 2D error\,$<$\,3\,px or relative error\,$<$\,5\%.
 As detailed in Suppl.~\ref{sec:comp_reas},
 all run times were measured on
 a single GPU (24GB NVIDIA Quadro RTX 6000).
 
\textbf{Methods.}
We study a wide range of methods and report the relevant metrics in Fig.~\ref{fig:kitti}: \\
In the \textbf{upper third} of the table, we report results
for unsupervised methods that do not require 
ground truth correspondences for training.
This includes the ``raw'' RobOT matching
of Eq.~(\ref{eq:robot_matching}), computed
on $(x,y,z)$ coordinates in $\mathbb{R}^3$
with a \emph{blur} scale $\sigma = 1$\,cm and a 
\emph{reach} scale $\tau = +\infty$.
Please also note that \text{PWC} refers to an improved version 
of \text{PointPWC-Net}, released on GitHub
~(\url{https://github.com/DylanWusee/PointPWC})
after the publication of \cite{wu2019pointpwc}
with a self-supervised loss.\\
In the \textbf{central third} of the table,
we benchmark a collection of state-of-the-art
point neural networks.
In the \textbf{lower third} of the table,
we study the influence of our RobOT-based layers.
The methods ``Pre + FLOT/PWC + Post''
correspond to the FLOT and PointPWC-Net architectures,
with the additional pre-alignment and post-processing modules
of Sec.~\ref{sec:reg_deep_deform}.
The last two lines correspond to the full
D-RobOT architecture (with a spline deformation model)
whose training is detailed in Suppl.~\ref{sec:suppl_kitti}.

\textbf{Results.}
We make three major observations:
\begin{compactenum}
\item Without any regularization or training,
a simple RobOT matching on high-resolution data
outperforms many deep learning methods in terms of speed,
memory footprint and accuracy (line 6 of the table).
This surprising result is strong evidence
that \textbf{geometric methods and baselines deserve more attention}
from the computer vision community.

\item In the lower third of the table, RobOT-enhanced methods
\textbf{consistently outperform state-of-the-art methods} by a wide margin.

\item As shown in Fig.~\ref{fig:kitti_visual_res}, these improvements are most significant on
\textbf{high-resolution data}.
\end{compactenum}

Overall, as detailed in Suppl.~\ref{sec:suppl_kitti}
and~\ref{sec:robot_vs_nn}, we observe that optimal transport theory
is especially well suited to scene flow estimation.
Assuming that ground points have been removed from the
3D frames (as a standard pre-processing), 
most object displacements can be explained as translations
and small rotations:
this is an ideal setting for our
robust geometric method.


\section{Registration of high-resolution lung vessel trees}
\label{subsec:deep_deformation_experiment}

\textbf{PVT1010: a new dataset for lung registration.}
Going further,
we introduce a new dataset of 1,010 pairs of 
pulmonary vessel trees
that must be registered between expiration (source) and inspiration (target).
Due to the intricate geometry of the lung vasculature 
and the complexity of the breathing motion, 
the registration of these shapes is a real challenge. 

As detailed in Suppl.~\ref{sec:suppl_lung_dataset},
we encode our 1,010\,$\times$\,2 vessel trees
as high-resolution 3D point clouds 
($\text{N}=\text{M}=$ 60k points per tree).
For each point, we also provide a local 
estimate of the vessel radius that we use
as an additional point feature
or as a weight $\alpha_i$ or $\beta_j$
in Eq.~(\ref{eq:unbalanced_Sinkhorn}).
Our first 1,000 pairs of 3D point clouds are provided
without ground truth correspondences;
for all of our experiments,
we randomly sample 600 training and 
100 validation cases from this 
large collection of unannotated patients.
The last 10 cases correspond to the 
10 \text{DirLab} \text{COPDGene} pairs \cite{castillo2013reference}:
they come with 300 expert-annotated 3D landmarks per lung pair,
that we use to test our methods.

\newcommand{\densepoints}[1]{\textcolor{red!60!black}{#1}}

\newcommand{\best}[1]{\textcolor{black}{\textbf{#1}}}
\newcommand{\bestdense}[1]{\textcolor{red!80!black}{\textbf{#1}}}

 \begin{figure}[p!]
\RawFloats

\includegraphics[width=\textwidth]{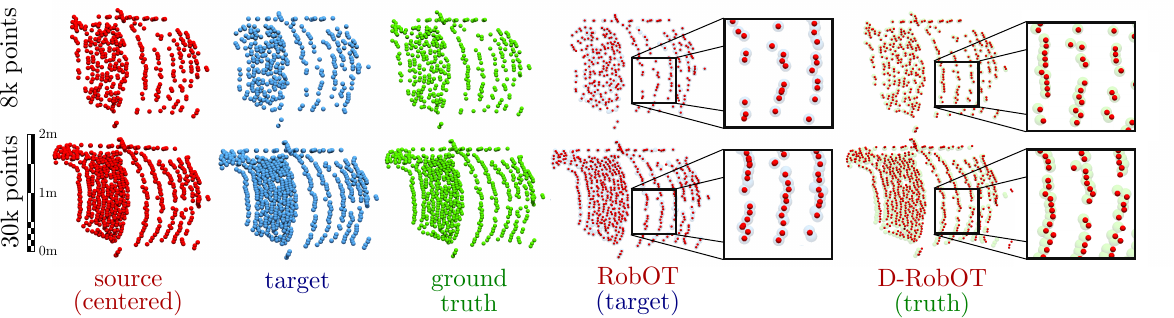}
\caption{\textbf{Influence of the sampling density} on a detail (top car) of the Kitti frames of
Fig.~\ref{fig:intro}.b.
\textbf{First row:} On sub-sampled 3D scenes, the regularizing priors
of the D-RobOT architecture prevent over-fitting to the random
sampling patterns of the target point cloud (\textcolor{blue!80!black}{blue}).
The D-RobOT output (last column, \textcolor{red!80!black}{red}) is very close to 
the ground truth scene flow (\textcolor{green!60!black}{green}).
\textbf{Second row:} Increasing the number of points per frame
reduces the influence of sampling artifacts.
The simple RobOT baseline (fourth column, \textcolor{red!80!black}{red}) 
still over-fits to the target (\textcolor{blue!80!black}{blue}) but becomes
remarkably competitive.}
\label{fig:kitti_visual_res}
\vspace*{.5cm}

  \begin{minipage}[b]{0.3\textwidth}
    \centering
    
\renewcommand{\arraystretch}{1}
\scalebox{0.5}{
\begin{tabular}{|p{0.6cm}p{3.6cm}crrcccccc|}
  \hline 
& \textbf{Method} 
& \textbf{Points}
& \textbf{Time}
& \textbf{Memory}
& \textbf{EPE3D} 
& \textbf{Acc3DS}
& \textbf{Acc3DR} 
& \textbf{Outliers3D}
& \textbf{EPE2D} 
& \textbf{Acc2D} \\
& 
&
& ms $\downarrow$
& Mb $\downarrow$~~ 
& cm $\downarrow$ 
& \% $\uparrow$ 
& \% $\uparrow$ 
& \% $\downarrow$ 
& px $\downarrow$ 
& \% $\uparrow$\\\hline 
\parbox[t]{1mm}{\multirow{6}{*}{\rotatebox[origin=c]{90}{\textbf{Unsupervised}}}} 
& \text{ICP} (rigid)~\cite{besl1992method}  
& ~~8k 
& 224~~
& \best{2}~~
&51.81 & ~~6.69 & 16.67 &87.12 &27.6752 &10.56\Tstrut\\
& FGR (rigid)~\cite{zhou2016fast}  
& ~~8k
& ----~~
& 30~~
&48.35 &13.31 &28.51 &77.61 &18.7464 &28.76\\
& \text{CPD} (non-rigid)~\cite{myronenko2010point} 
& ~~8k
& 34,880~~
& 798~~
& 41.44 &20.58 &40.01 &71.46 &27.0583 &19.80\\
& \text{PWC} (self)~\cite{wu2019pointpwc}  
& ~~8k
& 237~~
& 1,016~~
& 25.49 &23.79   & 49.57 &68.63 &~~8.9439    & 32.99\\
& \best{RobOT} (raw)
& ~~8k
& 170~~
& 3~~
&~~\best{9.12} & \best{60.43} & \best{79.39}  & \best{33.65} &~~\best{4.9920} &\best{56.23}\\
& \bestdense{RobOT}  \densepoints{(raw)}
& \densepoints{30k}
& \bestdense{166}~~
& \bestdense{89}~~
&~~\bestdense{4.67} &\bestdense{80.43} &\bestdense{91.05}  &\bestdense{20.21} & ~~\bestdense{1.7026} &\bestdense{85.71}\\

\hline 

\parbox[t]{1mm}{\multirow{12}{*}{\rotatebox[origin=c]{90}{\parbox{3cm}{\textbf{Supervised training }\\ \textbf{on FlyingThings3D}}}}} 
& \text{FlowNet3D}~\cite{liu2019flownet3d}   
& ~~8k
& ----~~
& 690~~
&17.67 &37.38 & 66.77 & 52.71 &~~7.2141 & 50.92\Tstrut\\
& \text{SPLATFlowNet}~\cite{su2018splatnet}   
& ~~8k
& ----~~
& ----~~
& 19.88 & 21.74 & 53.91 & 65.75 &~~8.2306 & 41.89\\
& original \text{BCL}~\cite{gu2019hplflownet}    
& ~~8k
& ----~~
& ----~~
& 17.29 & 25.16 & 60.11 & 62.15 &~~7.3476 & 44.11\\
& \text{HPLFlowNet}~\cite{gu2019hplflownet}   
& ~~8k
& ----~~
& ----~~
& 11.69 & 47.83 & 77.76 & 41.03 &~~4.8055 & 59.38\\
& \text{FLOT}~\cite{gu2019hplflownet}  
& ~~8k
& 324~~
& 2,826~~
&~~5.51 & 75.79 &  90.98 &  23.95 &~~3.3152 & 75.10 \\

& \text{PWC}~\cite{wu2019pointpwc}  
& ~~8k
& \best{237}~~
& 1,016~~
&~~5.28 &  85.83    & 94.08 & 18.85 &~~3.0074    &   81.48\\
& \densepoints{PWC} 
& \densepoints{30k}
& \densepoints{1,138}~~
& \densepoints{10,691}~~
&~~\densepoints{7.63} 
&  \densepoints{67.54} 
&  \densepoints{92.30} 
&  \densepoints{26.09} 
& ~~\densepoints{3.5212} 
&  \densepoints{70.74}\\\cline{2-11}
& Pre + FLOT + Post 
& ~~8k
& 487~~
& 2,826~~
& ~~5.33&  76.84 &   91.65 &  23.56 &~~3.2786 &  75.31\Tstrut\\
& Pre + PWC + Post 
& ~~8k
& 279~~
& 1,034~~
&~~3.35 &  90.10 &  97.32 &  \best{16.20} &~~\best{1.4301} &  \best{93.85}\\
& \densepoints{Pre + PWC + Post}
& \densepoints{30k}
& \densepoints{1,207}~~
& \densepoints{10,691}~~
& \densepoints{~~3.50} 
&  \densepoints{90.04} 
&  \densepoints{96.71} 
&  \densepoints{17.28} 
&~~\densepoints{1.5917} 
&  \densepoints{90.37}\\
& \best{D-RobOT} (spline)  
& ~~8k
& 268~~
& \best{396}~~
&~~\best{3.15} & \best{90.51}    &   \best{97.42} & 16.26 &~~1.4532    & 93.76\\
& \bestdense{D-RobOT} \densepoints{(spline)} 
& \densepoints{30k}
& \bestdense{547}~~
& \bestdense{610}~~
& ~~\bestdense{2.23} &  \bestdense{95.88} & \bestdense{99.19} &  \bestdense{12.89} &~~\bestdense{1.0336} & \bestdense{96.75}\\
\hline
\end{tabular}}

\end{minipage}

\vspace{-4.15cm}
\hspace{10.2cm}
  \begin{minipage}[t]{0.2\textwidth}
    \begin{tikzpicture}[scale=0.5]
    \begin{axis}[
        width=6.72cm, height=8.8cm,
        xmin=0, xmax=10,
        ymin=0, ymax=1250,
	    ytick={250,750},
	    extra x ticks={4,8},
	    extra y ticks={500,1000},
	    extra tick style={grid=major},
        xlabel={\textcolor{white}{.}~~~~~~~~~~~~~~~~EPE3D~(cm~$\downarrow$)},
        ylabel=Time~(ms~$\downarrow$)]
        \addplot[
            scatter/classes={a={mark=x,line width=1.5pt,blue}, b={mark=+,line width=1.5pt,red}},
            scatter, mark=*, only marks, 
            mark size=4pt,
            scatter src=explicit symbolic,
            nodes near coords*={\Label},
            visualization depends on={value \thisrow{label} \as \Label}, 
            visualization depends on={value \thisrow{anchor}\as\myanchor},
            every node near coord/.append style={anchor=\myanchor}
        ] table [meta=class] {
            x y class label anchor
            5.2825 237.0 a {~~PWC} west
            7.6313 1137.9 b {~~\densepoints{PWC}} west
            3.3479 278.9 a {Pre+PWC+Post~~~~~~~\textcolor{white}{.}\vphantom{$\int$}} south
            3.5004 1207.4 b {~~\densepoints{Pre+PWC+Post}} west
            3.1477 268.1 a {\best{D-RobOT}\vphantom{$\int$}} {north}
            2.2263 546.7 b {~~\bestdense{D-RobOT}} west
            9.1167 170.2 a {\best{RobOT}~~~~\textcolor{white}{.}\vphantom{$\int$}} {north}
            4.6743 165.9 b {\bestdense{RobOT}\vphantom{$\int$}} north
            5.5125 324.1 a {~~FLOT} west
        };
    \end{axis}
    \end{tikzpicture}
\end{minipage} 

\caption{\textbf{Evaluation on the \text{Kitti} dataset for 3D scene flow.} Black numbers and crosses ($\textcolor{blue}{\times}$) correspond
to results on scene pairs that are sampled with
8,192 points per frame;
\densepoints{red} numbers and plus signs ($\densepoints{+}$)
correspond to scene pairs that are sampled with 30,000 points per frame.}
\label{fig:kitti}

\vspace*{.5cm}

\includegraphics[width=.95\textwidth]{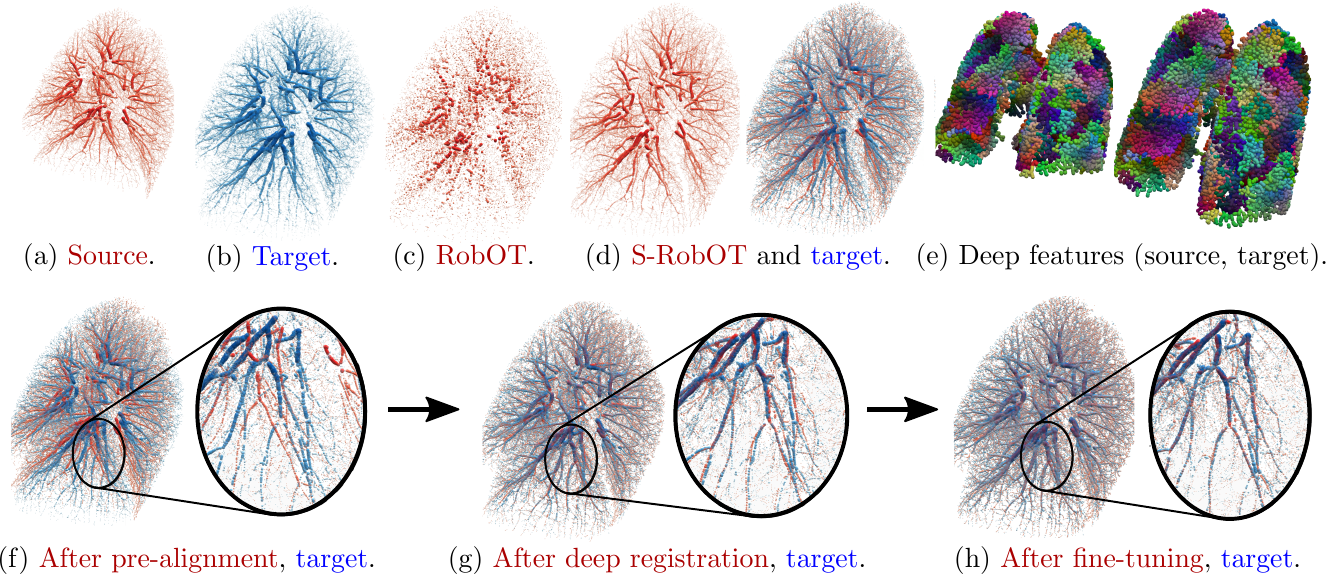}
\caption{\textbf{Registration of lung vascular trees.} 
\textbf{First row: S-RobOT} registration with the deep features
of Suppl.~\ref{sec:suppl_deep_feature_learning}. 
We display: (a) the source and (b) target shapes;
(c) the ``raw'' RobOT registration;
(d) a smoother S-RobOT registration with spline regularization,
an overlap of this result with the target shape;
(e) a visualization of the deep features
on a pair of lung vascular trees,
with colors that correspond to a 
t-SNE embedding of the point features in color space \cite{van2008visualizing}.
\textbf{Second row: D-RobOT} registration.
We display the successive steps of our model:
(f) pre-alignment with affine S-RobOT;
(g) deep registration with an LDDMM model;
(h) fine-tuning with spline S-RobOT.}
\label{fig:lung_visual_res}

\end{figure}
\clearpage

{\bf Data augmentation.}
The imbalance between our large training
set and the small collection of 10 test cases 
reveals a fundamental challenge
in computational anatomy:
annotating pairs of 3D medical shapes with 
dense correspondences
is prohibitively expensive.
To work around this problem,
we train our networks on synthetic deformations
of the 600\,$\times$\,2 lung shapes that make up our training set.
As detailed in Suppl.~\ref{sec:suppl_synth_data},
we use a two-scales random field to generate
a wide variety of deformations.
This allows us to
create a suitable training set 
with dense ``ground truth'' correspondences.

Overall, as discussed in Suppl.~\ref{sec:suppl_lung},
we found that supervised training on synthetic deformations
is both easier and more efficient than 
unsupervised training on real lung pairs.
From the chamfer and Wasserstein distances~\cite{feydy2020geometric} to local Laplacian penalties~\cite{wu2019pointpwc},
none of the unsupervised loss functions that we experimented with was able to deal with the complex geometry of our lung vascular trees. 
 
\textbf{Methods.}
We benchmark a wide range
of methods in Fig.~\ref{fig:lung_visual_res}
and Tab.~\ref{fig:dirlab}.
In the \textbf{upper half}
of the table, we evaluate geometric approaches that require no training;
in the \textbf{lower half} of the table, we benchmark scalable
point neural networks that we trained on our synthetic dataset.
We evaluate three types of RobOT-based approaches:
a simple RobOT matching computed using $(x,y,z)$ coordinates
as point features,
that may either be ``raw'' as in Eq.~(\ref{eq:robot_matching})
or regularized using
Eq.~(\ref{eq:affine});
an S-RobOT matching that we compute using
the deep features of Suppl.~\ref{sec:suppl_deep_feature_learning}
and regularize with the spline smoothing
of Eq.~(\ref{eq:spline})
or the LDDMM optimization of Eq.~(\ref{eq:shooting});
a D-RobOT architecture
that we pair with
three deformation models $\text{Morph}(\theta, x_i)\mapsto \hat{y_i}$
and describe in depth in Suppl.~\ref{sec:suppl_lung}.

\begin{table}[t!]

\centering

\hspace*{-.8cm}
\floatbox[{\capbeside\thisfloatsetup{capbesideposition={left,top},capbesidewidth=5cm}}]{table}[\FBwidth]
{\caption{\textbf{3D registration errors on the 3,000 expert-annotated DirLab landmarks.} 
The ${\dag}$ symbol denotes methods that are based on {\it image keypoints} \cite{hansen2019learning,hansen2020tackling} and are evaluated on the original DirLab \emph{image} dataset;
all other approaches are tested on our \emph{point clouds}. Due to its large memory requirements, \text{CPD} is tested on clouds of 20k points (instead of 60k).}
\label{fig:dirlab}}
{  \begin{minipage}[b]{0.5\textwidth}
    \centering
\scalebox{0.65}{
\renewcommand{\arraystretch}{1}
\begin{tabular}{|p{0.3cm}p{3.5cm}rrrrr|}
  \hline
& 
& \textbf{Average} 
& \multicolumn{3}{c}{\textbf{Percentiles}}
& \Tstrut\\
& & \textbf{error~~}
& \textbf{25\%~~} &  \textbf{50\%~~} & \textbf{75\%~~} 
& \textbf{Time} \\
& \textbf{Method} 
& mm $\downarrow$
& mm $\downarrow$
& mm $\downarrow$
& mm $\downarrow$
& s $\downarrow$ \\
\hline
\parbox[t]{1mm}{\multirow{5}{*}{\rotatebox[origin=c]{90}{\textbf{No training}}}} & \text{Input data} 
& 23.30 & 13.18 & 22.22 & 31.65 & ---~~\Tstrut\\
& \text{ICP} (affine) \cite{besl1992method} 
& 15.05 & 9.60 & 14.06 & 20.01 & 0.52\\
& \text{CPD} (non-rigid) \cite{myronenko2010point}
& 9.30 & 5.95 & 8.60 & 11.83  & 332.60\\
& \text{RobOT} (affine) 
& 10.45 & 6.01 & 9.83 & 13.97 & 0.18\\
& \text{RobOT} (raw)
& 9.41 & 4.89 & 8.35 & 13.04 & \textbf{0.15}\Bstrut\\
\hline
\parbox[t]{1mm}{\multirow{7}{*}{\rotatebox[origin=c]{90}{\textbf{Supervised}}}}
& \text{DGCNN}-\text{CPD}$^{\dag}$~\cite{hansen2019learning} &
4.30 & ---~~ & ---~~ & ---~~ & ---~~\Tstrut\\
& \text{DispEmd}$^{\dag}$~\cite{hansen2020tackling} & 3.42 & ---~~ & ---~~ & ---~~ & ---~~~\\
& \text{S-RobOT (spline)}
& 5.72 & 3.19 & 5.04 & 7.35 & 2.77\\
& \text{S-RobOT (LDDMM)}
& 5.48 & 2.86 & 4.44 & 7.14 & 42.30\\
& \text{D-RobOT} (raw)
& 3.40 & 1.40 & 2.58 & 3.69 & 1.26\\
& \text{D-RobOT} (spline)
& 2.95 & 1.30 & 2.50 & 3.19 & 1.87\\
& \text{D-RobOT} (\text{LDDMM})
& \textbf{2.86} & \textbf{1.25} & \textbf{2.23} & \textbf{3.11} &1.92\\


\hline
\end{tabular}
}

\end{minipage}}

\end{table}

{\bf Results.}
We provide additional experiments in Suppl.~\ref{sec:suppl_lung}
and make three major observations:
\begin{compactenum}
\item \textbf{Explicit regularization} with a spline or LDDMM deformation model is key.
Model-free architectures that predict raw 3D correspondences
produce non-smooth results
that are not anatomically plausible,
even when they are trained entirely on smooth deformations.
\item The D-RobOT architecture combines a \textbf{high acccuracy} with fast run times.
\item Most remaining errors occur at the \textbf{boundary of the lungs}, where acquisition artifacts prevent the
thinnest vessels from being sampled reliably in our point cloud representation.
\end{compactenum}

\section{Conclusion, limitations and future work}

Our work builds upon a decade of active research
in the field of computational optimal transport.
We leverage major advances on RobOT solvers
to define a new matching layer
which is a plug-and-play 
replacement for nearest neighbor projection.
This operation has two major uses in 3D shape registration:
first, it provides a \textbf{very strong
geometric baseline} for e.g. scene flow estimation;
second, it increases the accuracy
and generalization abilities of point neural networks
on finely sampled 3D shapes.
We see D-RobOT as a \textbf{mature and versatile architecture}
for shape registration which is easy to train
and adapt to task-specific requirements in e.g. medical imaging.

Going forward, we see three main ways of
improving this work.
First, we still have to investigate
in depth the important problem of occlusions and \textbf{partial
acquisitions}.
Second, integrating \textbf{task-specific features} 
beyond the $(x,y,z)$ point coordinates
is often key to perfect results.
In the specific setting of lung registration, 
working with image-based features or focusing
on branching points could be a way of improving
performance at the cost of portability:
recent works such as \cite{hansen2021deep,hansen2021graphregnet} are an excellent
source of inspiration.
Finally, we believe that high-quality \textbf{software packaging}
is an important part of research in our field.
We intend to keep working on the topic
and distribute our methods through a
user-friendly Python library for widespread use by
the scientific community.


\newpage

\begin{ack}
Research reported in this publication was supported by the National Heart, Lung, and Blood Institute of the National Institutes of Health under award numbers R01HL149877 and R01HL116473. The content is solely the responsibility of the authors and does not necessarily represent the official views of the National Institutes of Health.
The authors would also like to thank the anonymous
reviewers for their most valuable advice.
\end{ack}





\medskip

{\small
\bibliographystyle{abbrv}

}


\newpage 

\appendix
\section{Supplementary Material}
\label{sec:supplemental_material}

We provide additional information on our \text{RobOT} layer and the two derived methods:
\begin{itemize}
    \item \textbf{S-RobOT}: feature matching with \text{RobOT}, followed by a \textbf{smoothing} that is implemented in closed form or relies on an optimization loop -- as detailed in Sec.~\ref{sec:global_feature_matching}.
    \item \textbf{D-RobOT}: \text{S-RobOT} as a pre- and post-processing tool for a \textbf{deep} deformation estimator -- as detailed in Sec.~\ref{sec:reg_deep_deform}. 
\end{itemize}

More specifically:
\begin{compactenum}
\item Sec.~\ref{sec:suppl_lung_dataset} provides details on \textbf{PVT1010}, our dataset of pulmonary vascular trees. 
\item Sec.~\ref{sec:suppl_synth_data} describes our \textbf{synthetic deformations} for augmenting PVT1010.
\item Sec.~\ref{sec:suppl_fea_matching} contains more information on global feature matching with \textbf{S-RobOT}. 
\item Sec.~\ref{sec:suppl_deep_deform} provides details on our deep deformation prediction approach \textbf{D-RobOT}, with additional results on the PVT1010 and \text{Kitti} datasets. 
\item Sec.~\ref{sec:robot_vs_nn} discusses the differences between
    \textbf{RobOT and nearest neighbor projection}.
\item Sec.~\ref{sec:comp_reas} details our \textbf{computational resources}. 
\item Finally, Sec.~\ref{sec:supply_potential_for_negative} discusses the \textbf{societal impact} of our work.
\end{compactenum}

\subsection{Pulmonary vascular tree dataset}
\label{sec:suppl_lung_dataset}

\textbf{Introducing the \text{PVT1010} dataset.}
We introduce a new pulmonary vascular tree dataset for point cloud registration. The PVT1010 dataset includes 1,010 pairs of inhale/exhale lung vascular trees extracted from 3D computed tomography (CT) images; 10 of these correspond to the 10 cases of the public \text{DirLab}-\text{COPDGene}~\cite{castillo2013reference} dataset which includes, for each pair, 300 expert annotated landmarks that are in correspondence with each other and that we use to validate our results. We extracted the lung vascular geometry in both inspiratory and expiratory CT scans using a scale-space particle system ~\cite{kindlmann2009sampling,estepar2012computational} that is implemented in the \text{Teem} library~\cite{teem}. We used the pipeline that is defined in the chest imaging platform~\cite{nardelli2020generative,chestimgplat}. 

\textbf{Legal and regulatory information.}
The vascular tree reconstructions that are used in this study were part of the COPDGene study (NCT00608764). This study has been IRB approved and participants have provided their consent. The investigators from the Brigham and Women's Hospital (Harvard Medical School) only had access to de-identified CT images to perform vascular reconstructions. Since we are performing secondary analysis using de-identified data, the work under consideration is not considered human subjects research and did not imply additional risks to the participants. Risks related to ionizing radiation exposure were described in the primary IRB-approved study. Study identifiers were re-coded for our release of \text{PVT1010} to preserve anonymity. \text{PVT1010} is released under the {\it Creative Commons Attribution-NonCommercial-ShareAlike 3.0 License}.

\textbf{Point cloud representation.}
We now detail how we extracted the lung vascular trees as high-resolution 3D point clouds from the raw CT images. To perform this geometric segmentation task, we rely on a system of 4D particles that are defined by three spatial coordinates plus one scale parameter that corresponds to the local radius of the lung vessel -- these radii are used by \text{RobOT} as point weights $\alpha_i$ and $\beta_j$ in Eq.~(\ref{eq:unbalanced_Sinkhorn}). 
Our segmentation method starts from a point cloud 
that is initialized using the Frangi filter \cite{frangi1998multiscale}.
Then, we fit this point cloud representation to our 3D CT volumes by minimizing iteratively a system energy that is expressed as the sum of:
\begin{itemize}
    \item An \textbf{inter-particle regularization} energy that ensures convenient sampling properties. We rely on a sum of quartic polynomials of the pairwise point distances, with a tunable potential well that is chosen to induce regular sampling at a fixed distance between the points.
    \item A \textbf{particle-image data fidelity} term, which is computed using the image Hessian at the current particles’ locations. 
\end{itemize}
As a final result, we obtain points that are approximately equally distributed along the vessel centerlines. The vessel radii are first approximated as the image scales at which the middle eigenvalues of the Hessian are locally minimized,
and then refined using the generative approach of~\cite{nardelli2020generative}. 

The resulting dataset has the following properties, which result in a challenging registration task: 
\begin{enumerate}
    \item The vascular trees have a \textbf{complex structure} and exhibit \textbf{complex, large motions} between inhalation and exhalation.
    \item To capture the complex anatomical structure of the lungs at millimeter scale, registration methods need to focus on branching points or rely on \textbf{high-resolution} point clouds.
    \item  Due to the fixed image resolution of the raw CT volume and to acquisition differences between the inhale and exhale scans, the extracted inhale and exhale vascular trees are \textbf{not fully consistent} with each other. This is especially true for tiny structures at the lung boundary, that may not be visible on the smaller lung images at exhalation time.
\end{enumerate}

\textbf{Sampling density, resolution.}
The original CT images from which we extract our point clouds
are acquired with a uniform resolution on the
$x$, $y$ and $z$ axes:
depending on the patients, the side length
of our voxels varies between 0.60\,mm and 0.65\,mm.
Using the processing above, we turn these volumetric
images into 3D point clouds with 60k samples
per lung vascular tree:
depending on the subject, the average sampling distance 
(from each point to its nearest neighbor in the 3D point cloud)
ranges between 0.6\,mm and 1.0\,mm.

We note that for our 10 test cases, the Dirlab annotations
were performed on a down-sampled volume 
with a resolution of 2.5\,mm on the $z$ axis.
To guarantee a fair comparison between
image-based and point-based methods,
we follow standard practice for this dataset
(\url{https://www.dir-lab.com/Results.html})
and report our results in Tab.~\ref{fig:dirlab}
with a ``snap-to-voxel'' post-processing:
we quantize our 3D lung registrations on the
original grids (with spacing $\sim$ 0.625\,mm\,$\times$\,0.625\,mm\,$\times$\,2.5\,mm)
before computing the average errors and percentiles.
In practice, we note that this quantization slightly
lowers the 25\% percentile of the registration errors
(as we ``snap'' many displacements to the correct voxel
or slice)
but increases the 50\% and 75\% percentiles
(some landmarks get ``snapped'' to the wrong slice).
Please note that in the Supplementary Material,
we do not include this quantization step
for e.g. ablation studies:
this allows us to study more precisely
the impact of each layer in our architecture.

\textbf{Volume vs point cloud representation.}
We stress that our point clouds contain much less information
than the original 3D volumes from which they have been sampled.
We discard all the intensity (grayscale) values
and only retain the sparse geometric support of the
lung vascular tree -- 60k points out of 100M+ voxels.
As a consequence, our registration task on 
3D point clouds is significantly harder than 
the original DirLab benchmark (\url{https://www.dir-lab.com/Results.html}).
Whereas optimization-based methods on the full
CT \textbf{volumes} reach a nearly perfect accuracy of 
0.60\,mm to 1.00\,mm with run times on the order of the \textbf{minute}
\cite{castillo2013reference,ruhaak2013highly,polzin2013combining,castillo2014computing,hermann2014evaluation,heinrich2015estimating,polzin2016memory,vishnevskiy2016isotropic,ruhaak2017estimation,vishnevskiy2017isotropic},
our \textbf{point} neural networks reach an average accuracy
of 2\,mm to 4\,mm in one or two \textbf{seconds}.

The main purpose of our work on the PVT1010 dataset
is to show that fast and accurate registration
is now at hand, even on very degraded anatomical data.
This is of significant interest for clinical practice:
our method is suitable for \textbf{real-time processing}
and has \textbf{built-in robustness}
to changes of the CT acquisition parameters
that may affect the intensities of the raw image volumes.
Going forward, as detailed in the conclusion of our manuscript,
we intend to work on improving the accuracy
of our method with a better sampling strategy
and image-based features.
Packaging our method as an accessible Python toolbox
will also open the door to a genuine 
multi-center evaluation of our trained models.

\subsection{Augmentation of the training dataset for vascular tree registration}
\label{sec:suppl_synth_data}

We now describe how to augment the \text{PVT1010} dataset
with synthetic deformations in order to create
a large training set with \textbf{dense ground truth annotations} for lung registration. We proceed in four steps: voxel-grid sampling; local deformation; global deformation; local property distortion and degradation using an inconsistent sub-sampling. Our efficient implementation lets us generate synthetic training pairs online -- just like a standard data augmentation layer. We showcase our local and global deformations in Fig.~\ref{fig:synthesize_data}.

{\bf 1. Voxel-grid sampling.} To start, we use a voxel-grid strategy to re-sample the raw point clouds with a standard sampling density. First, we subdivide the volume space into coarse 3D blocks with spacing $s_{\text{voxelgrid}} = 0.03$\,mm.
Second, we sort all points into these cubic bins according to their $(x, y, z)$ coordinates. Third, we compute one barycenter per cubic cell to down-sample the original point cloud. Since we work with \emph{weighted} point clouds, these local centroids are associated to the sums of the weights of all points in the corresponding cells.

{\bf 2. Local deformation.} We then sample control points from the vessel trees and generate random displacements that we smooth using an anisotropic spline model:
\begin{enumerate}
    \item We first sample $\text{C} = 1,000$ spline control points $x_c$ uniformly at random from the point cloud $(x_1, \dots, x_\text{N})$.
    \item Second, we compute local covariance matrices for the distribution of points $x_i$ in a neighborhood of each control point $x_c$, using an isotropic Gaussian kernel window.
    \item We compute the three eigenvalues $(e_c^1, e_c^2, e_c^3)$ and unit eigenvectors $(v_c^1, v_c^2, v_c^3)$ of each local covariance matrix to determine the main direction of the lung vessel.
    To avoid rank deficiency, we use a lower threshold of $0.2$ on the eigenvalues $e_c^k$. For each control point $x_c$, we then normalize the vector of three eigenvalues $(e_c^1, e_c^2, e_c^3)$ as $(e_c^1, e_c^2, e_c^3) / \sqrt{(e_c^1)^2 + (e_c^2)^2 + (e_c^3)^2 }$.
    \item For each control point $x_c$, we create a numerically stable anisotropic covariance matrix $\Sigma_c =\sum_{k=1}^3 (s_{\text{local}}e_c^k)^2 \, {v_c^k}{v_c^k}^\top$, where $s_{\text{local}} = 4$\,mm is a positive scaling factor.
    \item To obtain a robust estimation of the local covariance structure of our point cloud, we re-run steps 2-4 with neighborhoods that are defined using an \emph{anisotropic} Gaussian kernel window of covariance $\Sigma_c$.
    \item For every control point $x_c$, we generate a random displacement vector
    $\Delta x_c\in \mathbb{R}^3$ such that $\|\Delta {x_c}\| \leqslant d_{\text{local}}$, drawn uniformly in the ball of center $0$ and radius $d_{\text{local}} = 3$\,mm.
    \item We smooth and interpolate the displacement vector field $\Delta {x_c}$ from 
    the control points $x_c$ to the full point cloud $\{ x_i \}$ using an anisotropic Nadaraya--Watson kernel interpolator:
    \begin{align}
        x_i ~\gets~x_i ~+~ \frac{\sum_{c=1}^\text{C} k_{\Sigma_c}(x_c,x_i)\Delta {x_c}}{\sum_{c=1}^\text{C} k_{\Sigma_c}(x_c,x_i)}~,
    \end{align}
    where $k_{\Sigma_c}(x_c,\cdot)$ is a Gaussian kernel with local covariance $\Sigma_c$.
    This anisotropic formula ensures that the \textbf{local connectivity structure} of the lung vessel tree is preserved: the displacement of points that belong to the same vessel are strongly correlated with each other.
\end{enumerate}

\afterpage{
\begin{figure}[p]
\RawFloats
\includegraphics[width=1\textwidth]{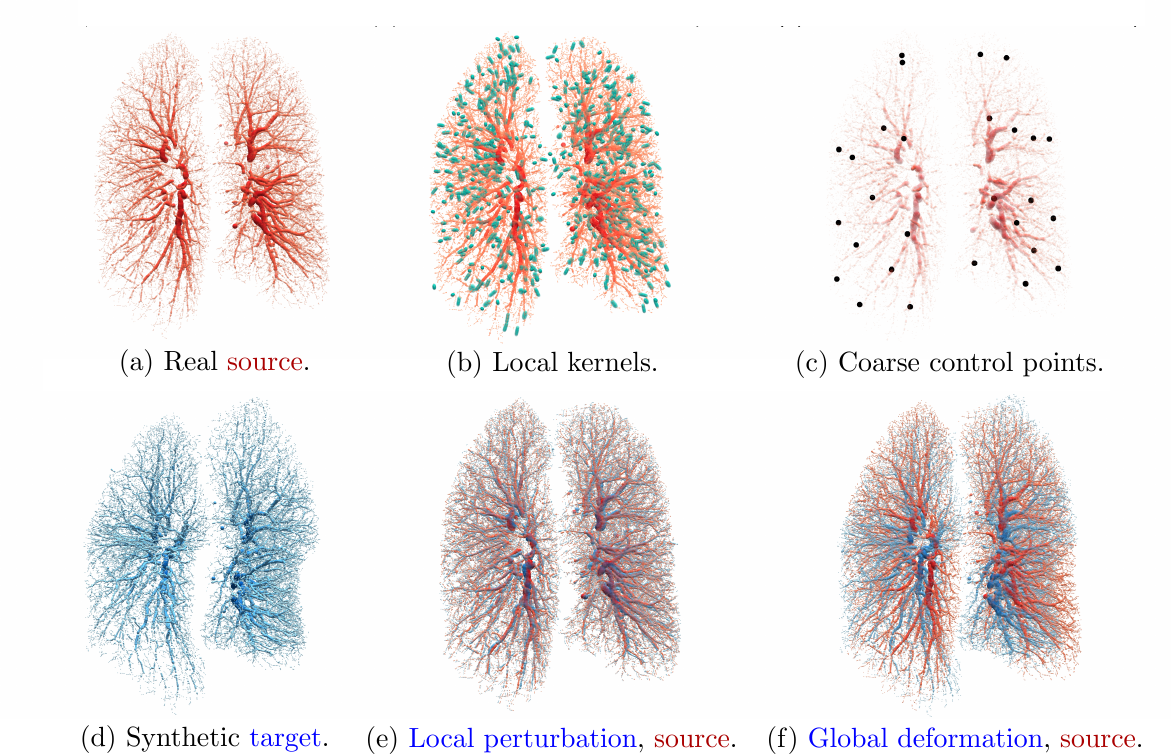}
\caption{\textbf{Local and global deformations that we use to generate our synthetic training dataset.} (a)~Original (real) vascular tree. (b)~Ellipsoids that represent anisotropic kernels of size 2\,mm that we use to generate vessel-preserving local deformations. Note that in our experiments, we use larger kernels of size 4\,mm
that induce a stronger regularization but are harder to display cleanly. (c)~Spline control points that we sample using a voxel-grid scheme with 90\,mm spacing and use to generate a global deformation.
(d)~Synthetic vascular tree, the output of the process that we use as a target for training. (e)~Source point cloud after local deformation. (f)~After global deformation.}
\label{fig:synthesize_data}

\vspace*{.5cm}

\centering
\includegraphics[width=.9\textwidth]{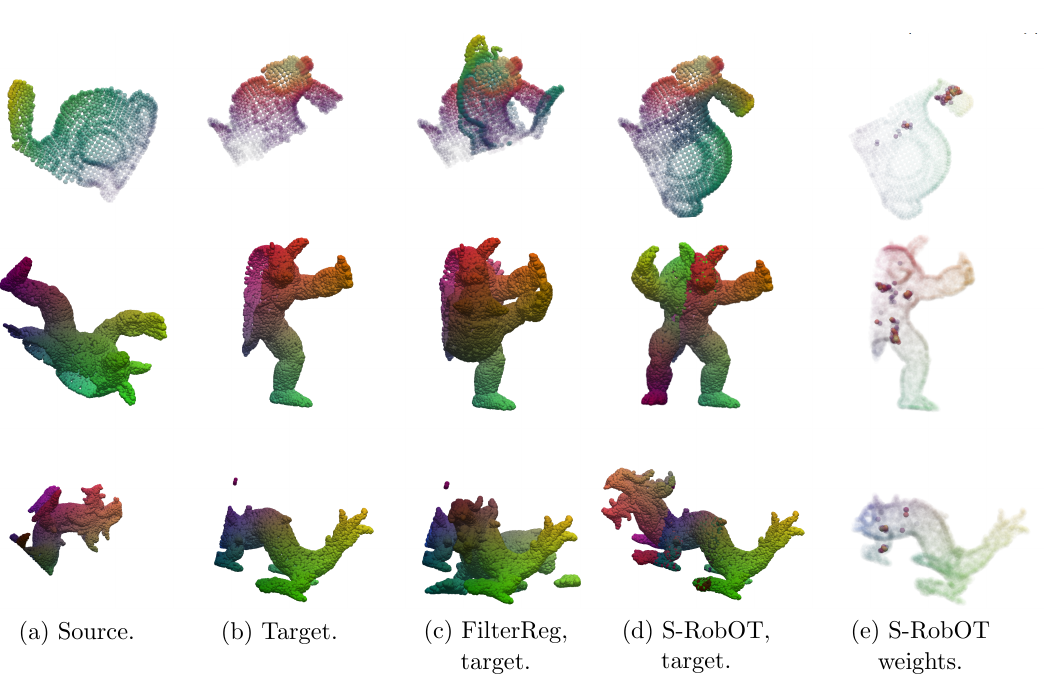}
\caption{\textbf{Partial rigid registration} of the Stanford scans based on the matching of \text{FPFH} features -- as discussed in Sec.~\ref{sec:global_feature_matching} and Suppl.~\ref{sec:suppl_partial_matching}. 
The feature-based \text{CPD} model \text{FilterReg} 
falls in a local minimum while the
rigid \text{S-RobOT} of Eq.~\eqref{eq:rigid} results in a successful registration. In the last column, we display the attention weights $w_i$ that are derived from unbalanced OT.}
\label{fig:partial_rigid}

\end{figure}

\begin{figure}[p]
\RawFloats

\includegraphics[width=1\textwidth]{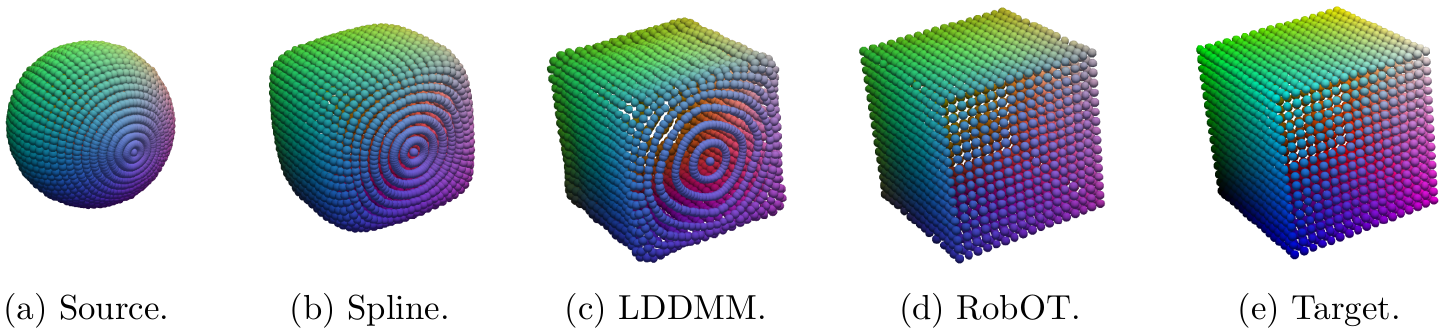}
\caption{\textbf{Smooth \text{RobOT} (\text{S-RobOT}) on a toy registration problem,} the deformation of a sphere (left) onto a cube (right). 
From left to right, we display:
(a) the source shape $\text{A} = (x_1, \dots, x_\text{N})$;
(b) the output of Spline RobOT from Eq.~(\ref{eq:spline});
(c) the output of LDDMM RobOT, solution of the optimization
problems of Eq.~(\ref{eq:morph_optimization},\ref{eq:shooting});
(d) the output of the ``raw'' RobOT matching
$x_i \mapsto x_i + v_i$
from Eq.~(\ref{eq:robot_matching});
(e) the target shape $\text{B} = (y_1, \dots, y_\text{M})$.}
\label{fig:reg_regress}

\vspace*{.5cm}

\end{figure}

\begin{table}[p]
\RawFloats

\scalebox{.95}{
\begin{tabular}{|llrrrrrr|}
  \hline
  & &
  \multicolumn{3}{c}{\textbf{Rotation}}&
  \multicolumn{3}{c}{\textbf{Translation}} \\
 && \textbf{MSE}~~~ & \textbf{RMSE}~~ & \textbf{MAE}~~& \textbf{MSE}~~~ & \textbf{RMSE}~~ & \textbf{MAE~~}  \\
 &\textbf{Method}& degrees${}^2$ $\downarrow$ 
 & degrees $\downarrow$ 
 & degrees $\downarrow$
 & 3D  units${}^2$ $\downarrow$   
 & 3D units $\downarrow$  
 & 3D units $\downarrow$ \Bstrut \\
\hline
\parbox[t]{1mm}{\multirow{4}{*}{\rotatebox[origin=c]{90}{\scriptsize{\textbf{Matching}}}}}
& ICP &
1134.552 & 33.683 & 25.045 &0.0856 &0.293 &0.250\Tstrut\\ 
&FGR~\cite{zhou2016fast}  & 
126.288  & 11.238 &2.832 &0.0009  &0.030 &0.008 \\ 
&Go-ICP~\cite{yang2015go} & 195.985  & 13.999  & 3.165  &0.0011  & 0.033 &0.012\\
&S-RobOT (rigid) & 8.939 & 2.989 & 1.313  & 0.0002  &0.014  &0.009\\
\hline
\parbox[t]{1mm}{\multirow{4}{*}{\rotatebox[origin=c]{90}{\scriptsize{\textbf{End-to-end}}}}}
&PointNetLK~\cite{aoki2019pointnetlk} &
 280.044 &16.735 &7.550 & 0.0020 &0.045  & 0.025\Tstrut \\ 
&DCP(v2)~\cite{wang2019deep} &
45.005  & 6.709  &4.448  &0.0007  & 0.027 &0.020\\
&PRNet~\cite{wang2019prnet} &
10.235  & 3.199 &1.454  & 0.0003 &0.016 & 0.010\\
&Partial-OT~\cite{dang2020learning} &
\textbf{0.107}  & \textbf{0.328}  &
\textbf{0.052}  & \textbf{3.384e-06}  & 
\textbf{0.00183}  & \textbf{0.0003}\Bstrut\\
\hline 
\end{tabular}}
\caption{\bf{Partial-to-Partial registration with unseen objects on ModelNet40}. }
\label{tab:seen_modelnet}

\vspace*{.5cm}

\scalebox{.95}{
\begin{tabular}{|llrrrrrr|}
  \hline
  & &
  \multicolumn{3}{c}{\textbf{Rotation}}&
  \multicolumn{3}{c}{\textbf{Translation}} \\
 && \textbf{MSE}~~~ & \textbf{RMSE}~~ & \textbf{MAE}~~& \textbf{MSE}~~~ & \textbf{RMSE}~~ & \textbf{MAE~~}  \\
 &\textbf{Method}& degrees${}^2$ $\downarrow$ 
 & degrees $\downarrow$ 
 & degrees $\downarrow$
 & 3D  units${}^2$ $\downarrow$   
 & 3D units $\downarrow$  
 & 3D units $\downarrow$ \Bstrut \\
\hline
\parbox[t]{1mm}{\multirow{4}{*}{\rotatebox[origin=c]{90}{\scriptsize{\textbf{Matching}}}}}
&ICP & 
1217.618 & 34.894 & 25.455 &0.086 &0.293 &0.251\Tstrut \\ 
&FGR~\cite{zhou2016fast}  & 
98.635 & 9.932 &1.952 &0.0014 &0.038 &0.007 \\ 
&Go-ICP~\cite{yang2015go} & 157.072  & 12.533  & 2.940 &0.0009  & 0.031 &0.010\\
&S-RobOT(rigid) &50.997 & 7.142
  &4.012  & 0.0005  &0.022  &0.013\\
\hline
\parbox[t]{1mm}{\multirow{4}{*}{\rotatebox[origin=c]{90}{\scriptsize{\textbf{End-to-end}}}}}
&PointNetLK~\cite{aoki2019pointnetlk} &
 526.401 &22.943 &9.655 & 0.0037&0.061 & 0.033\Tstrut\\ 
&DCP(v2)~\cite{wang2019deep} &
95.431  & 9.769  &6.954  &0.0010  & 0.034 &0.025\\
&PRNet~\cite{wang2019prnet} &
24.857  & 4.986  &2.329  &0.0004   & 0.021 &0.015\\
&Partial-OT~\cite{dang2020learning} &
\textbf{0.127}  &\textbf{0.357}  &\textbf{0.069}  &\textbf{3.953e-06}  &\textbf{0.002}  &\textbf{0.0004}\Bstrut\\
\hline 
\end{tabular}}
\caption{\bf{Partial-to-Partial registration with unseen categories on ModelNet40}. }
\label{tab:unseen_modelnet}
\end{table} 
\clearpage}

{\bf 3. Global deformation.} The step above simulates local relative displacements between lung vessels. To take large-scale breathing movements into account, we apply a second spline deformation which is smoother but has a larger magnitude.
In practice, we use a voxel-grid sampling with spacing resolution $s_{\text{global}} = 90$\,mm to obtain control points. We then generate random displacements of magnitude at most $d_\text{global}$ for every control point, and interpolate them to the full point cloud using a Nadaraya--Watson estimator, parameterized by an isotropic Gaussian kernel with standard deviation $\sigma_{\text{global}}$.

{\bf 4.a. Radius distortion.} Having altered the 3D coordinates of our points using the deformations above, we add random noise to the local estimates $\alpha_i$ of the vessel radii -- which are encoded as additional point features as detailed in Sec.~\ref{sec:suppl_lung_dataset} and used in our \text{RobOT} layer as point weights. This additive noise is scaled by a positive parameter $s_{\text{radius}} = 0.1$ and drawn at random in $[-s_{\text{radius}}\alpha_i, s_{\text{radius}}\alpha_i]$. 

{\bf 4.b. Inconsistent sampling.} By construction, the steps above let us create an arbitrary number of pairs of (real, simulated) lungs with known pairwise correspondence between all points. In order to simulate acquisition artifacts and introduce challenging inconsistencies, we sample $\text{N} = \text{M} = 60$k points at random from the source and the synthetic (target) point clouds as a last generation step. We note that for training,
the ground-truth flow is computed based on the source sampling and is not affected by this last degradation:
it may or may not point to a sample in the synthetic target.

{\bf Augmentation of the the target and source point clouds.} In our experiments, we generate our target point clouds using $d_{\text{global}} = 25$\,mm, $\sigma_{\text{global}} = 25$\,mm. Additionally, we also perform data augmentation for the source point cloud itself using the smaller values of $d_{\text{global}} = 8$\,mm and $\sigma_{\text{global}}  = 15$\,mm.

\newpage 

\subsection{Additional material on S-RobOT and global feature matching}
\label{sec:suppl_fea_matching}

\subsubsection{Partial registration} 
\label{sec:suppl_partial_matching}

{\bf Experiment 1: Stanford scans.}
We now discuss a toy example, illustrated in Fig.~\ref{fig:partial_rigid}, that showcases the benefits of unbalanced optimal transport theory for partial rigid registration. 
We normalize and resample the partial Stanford scans using a voxel-grid sampling with 0.005 spacing. We then rotate every source shape by (120°,10°,10°) degrees to create a target shape. 
We illustrate two methods:
\begin{itemize}
    \item As a baseline competitor, we use the \text{FilterReg}~\cite{gao2019filterreg} implementation of \url{https://github.com/neka-nat/probreg} (a feature-based \text{CPD} method \cite{myronenko2010point}) with noise ratio set to 0.7 and an automatic annealing strategy based on the Expectation-Maximization (EM) algorithm. 
    \item For our rigid \text{S-RobOT} registration, we first compute a 33-dimensional \text{FPFH} feature vector \cite{rusu2009fast} for each point using a radius of 0.02 for the normal search and a radius of 0.05 for the feature search. We then rely on Eqs.~(\ref{eq:robot_matching}-\ref{eq:rigid})
    to compute the weighted RobOT matching and project it onto the space
    of rigid transformations. For our robust OT problem,
    we set the \emph{blur} parameter to 0.1 and the \emph{reach} parameter to 2 units in $\mathbb{R}^{33}$.
\end{itemize}

{\bf Results.}
Given standard \text{FPFH} features \cite{rusu2009fast}, our \text{RobOT} approach successfully registers the low-overlap pair while \text{FilterReg}~\cite{gao2019filterreg} fails. The \text{RobOT} confidence weights $w_i$ of Eq.~\eqref{eq:robot_matching} naturally act as an attention mechanism.

\textbf{Experiment 2: ModelNet40.}
Going further,
we evaluate our partial registration strategy S-RobOT on
the standard dataset ModelNet-40~\cite{wu20153d}
with comparisons to state-of-the-art methods. 
Instead of relying on FPFH features, 
we use a self-supervised deep feature learning strategy
to increase the robustness and accuracy of the registration. 
The ModelNet40 dataset contains 12,311 CAD (Computer-Aided Design) models,
from 40 object categories. 
We follow the same experimental setup as
in~\cite{wang2019prnet,dang2020learning}:
\begin{enumerate}
    \item We normalize the point clouds to fit in the cube $[-1,1]^3$.
    \item To generate a registration pair, we first sample a random source shape
       using 1,024 points. We then apply a random rigid
       transformation along each axis, with rotation angle in $[0^{\circ},45^{\circ}]$ and translation in $[-0.5,+0.5]$.
    \item To simulate a partial acquisition, we sample one point at random
    in both of the source and target shapes. In each shape,
    we then keep the 768 nearest neighbors of these points
    to define the observed regions.
\end{enumerate}

\textbf{Evaluation metrics.}
To assess the quality of our rigid registrations,
we evaluate the rotation and translation errors separately.
On each component, we compute the mean squared error (MSE), 
the root mean squared error (RMSE)
and the mean absolute error (MAE).

\textbf{S-RobOT setup.} For feature learning, we proceed in two steps: 
\begin{enumerate}
    \item Given a source point cloud, we synthesize a target point cloud by applying a random rigid transform. For convenience, we re-use the augmentation strategy detailed above: along each axis, we sample a rotation in $[0^{\circ},45^{\circ}]$ and a translation in $[-0.5,+0.5]$.
    \item We train a \textbf{self-supervised deep feature extractor} using the
    loss function of Sec.~\ref{sec:suppl_deep_feature_learning}. 
    We use a PointNet++~\cite{qi2017pointnet++} with 10,654 parameters 
    that learns a 30-dimensional feature vector for each point.
\end{enumerate}
For the S-RobOT matching, we rely on the rigid projection formula of Eq.~(\ref{eq:rigid}); we set the \emph{blur} parameter to $\sigma = 0.01$ and the \emph{reach} parameter to $\tau = 10$ units in $\mathbb{R}^{30}$.

\textbf{Partial-to-Partial registration with unseen objects.} 
We follow~\cite{wang2019prnet,dang2020learning}
and split the dataset of 12,311 point clouds into a training set
with 9,843 shapes and a testing set with 2,468 shapes.
We first train on \textbf{all} 9,843 shapes from 
\textbf{all 40 categories} in the training set and 
test on \textbf{all} 2,468 unseen shapes in the test set. 
In Table~\ref{tab:seen_modelnet}, we compare S-RoboT with feature matching methods (ICP, FGR~\cite{zhou2016fast}, Go-ICP~\cite{yang2015go}) and end-to-end deep learning models (PointNetLK~\cite{aoki2019pointnetlk}, DCP(v2)~\cite{wang2019deep}, PRNet~\cite{wang2019prnet} and Partial-OT~\cite{dang2020learning}).

\textbf{Partial-to-Partial registration with unseen categories}. 
We then follow~\cite{wang2019prnet} and test the generalization ability of our model between object categories:
we train on the first 20 categories of ModelNet40 and test on the remaining 20 categories. We report these results in Tab.~\ref{tab:unseen_modelnet}.

\textbf{Results.} 
Overall, we observe that the unsupervised 
Rigid S-RobOT method performs much better 
than traditional approaches and is close to end-to-end methods. 
Since our main focus is on free-form registration,
we do not push these experiments further:
we use Affine and Rigid S-RobOT as pre-alignment
steps and are satisfied with this level of performance.

We note that the best-performing method Partial-OT 
also relies on an optimal transport layer: 
as discussed in \cite{sejourne2019sinkhorn},
\textbf{partial} OT is a very close cousin of the theory of
\textbf{unbalanced} OT that we leverage in our RobOT layer. 
We understand the Partial-OT method as a supervised
and end-to-end version of our S-RobOT baseline. 
Both~\cite{dang2020learning}
and this manuscript thus show that
\textbf{optimal transport theory is ready to be part
of the standard toolbox in our field},
with state-of-the-art performance in
varied and complementary application settings.

\subsubsection{Diffeomorphic registration} 
\label{sec:suppl_lddmm}

\textbf{Background on LDDMM.} 
Going beyond rigid and affine transformation models, the Large Deformation Diffeomorphic Metric Mapping (\text{LDDMM}) framework captures large, smooth and invertible deformations in a principled way~\cite{beg2005computing}. This fluid-based model is standard in computational anatomy, especially for applications to neuroimaging where the preservation of shape topology is a key registration prior~\cite{feydy2020geometric}.

In the \text{LDDMM} model, deformations are encoded via the integration of a time-varying velocity field $\textit{v}^t =\frac{\mathrm{d}}{\mathrm{d} t} x^{t}$ over the unit time interval $t \in [0,1]$. Given any two shapes A and B to register with each other, we seek a geodesic path $(\textit{v}^t)_{t\in[0,1]}$ for a chosen Riemannian metric $\|\textit{v}\|_{K}^{2}$ that is induced by a convolution kernel $K_{x}$ over the space of vector fields $\textit{v}:\mathbb{R}^3\rightarrow \mathbb{R}^3$. 

Following standard derivations from optimal control theory~\cite{beg2005computing,hart2009optimal,feydy2020geometric},
we know that plausible deformations are fully parameterized by the \textbf{momentum} $m^0 : \mathbb{R}^3\rightarrow \mathbb{R}^3$
at time $t=0$, a vector field that acts as the parameter ``$\theta$'' of the \text{LDDMM} deformation model. 
In the LDDMM framework, the optimization problem
of Eq.~(\ref{eq:morph_optimization}) reads:
\begin{equation}\label{eq:shooting}
\theta^* = m^{0*} = \argmin_{m^0:\mathbb{R}^3\rightarrow\mathbb{R}^3} ~
\underbrace{\vphantom{\sum_{i=1}^\text{N}}\lambda_\text{reg}\,\langle m^0, K_x*m^0\rangle_{L^2(\mathbb{R}^3, \mathbb{R}^3)}}_{\text{Regularization.}}  
+  
\underbrace{\sum_{i=1}^\text{N} w_i \|x_i + v_i - \text{Morph}(m^0, x_i)\|_{\mathbb{R}^3}^2}_{\text{Fidelity to the data.}}~, 
\end{equation}
where the deformation model $\text{Morph} : (m^0, x^0_i) \mapsto x^1_i$ is computed through the integration of the geodesic shooting equation:
\begin{equation}
~ \tfrac{d}{d t} x^{t}=+\tfrac{\partial H}{\partial m}\left(x^{t}, m^{t}\right),~\tfrac{d}{d t} m^{t}=-\tfrac{\partial H}{\partial x}\left(x^{t}, m^{t}\right)\, \label{eq:lddmm_hamiltonian}
\end{equation}
from time $t=0$ to time $t=1$
for the Hamiltonian $H(x, m) = \tfrac{1}{2}\langle m, K_x * m \rangle_{L^2(\mathbb{R}^3, \mathbb{R}^3)}$.
We refer to Chapter~5 of~\cite{feydy2020geometric} for a
presentation and implementation of these equations
on point clouds with the
\text{PyTorch} and \text{KeOps} libraries.

{\bf Black-box deformation models.}
Different transformation models may result in different projection results. In Fig.~\ref{fig:reg_regress}, we show several smooth \text{RobOT} (\text{S-RobOT}) results for spline and \text{LDDMM} models on a toy registration task.

\textbf{Experiment.} The source sphere (left) and the target cube (right) are sampled with 
$\text{N} = 1,922$ and 
$\text{M} = 1,538$ points respectively. We normalize their coordinates to be contained in $[-1,1]^3$ and normalize the point cloud weights to sum up to one ($\alpha_i = 1/1,922$ and $\beta_j = 1 / 1,538$ so that $\sum_{i=1}^{1,922}\alpha_i = \sum_{j=1}^{1,538} \beta_j= 1$).  For the initial \text{RobOT} matching, we set the \emph{blur} parameter to 0.005 units in $\mathbb{R}^3$ and the \emph{reach} parameter to $+\infty$ (balanced OT). For the following spline and \text{LDDMM} regularizations, we use a multi-Gaussian-kernel with standard deviations $\{0.05,0.2,0.3\}$ and weights $\{0.2,0.3,0.5\}$, i.e. a kernel function:
\begin{align}
    k(x,y) ~&=~ 
    0.2 \cdot \exp\big[ -\|x-y\|^2_{\mathbb{R}^3} / (2 \cdot 0.05^2)\big]
    ~+~
    0.3 \cdot \exp\big[-\|x-y\|^2_{\mathbb{R}^3} / (2 \cdot 0.2^2)\big] \\
    ~&+~
    0.5 \cdot \exp\big[-\|x-y\|^2_{\mathbb{R}^3} / (2 \cdot 0.3^2)\big]~. \nonumber
\end{align}

{\bf Results.}
We make two important observations:
\begin{itemize}
\item As evidenced by the disappearance of the 
\textbf{polar sampling pattern} in the fourth column,
the raw \text{RobOT} matching $x_i \mapsto x_i+v_i$ provides a ``perfect fit'' to the target but does not preserve the topology of the source point cloud. 
The spline and the \text{LDDMM} approximations alleviate this problem: they smooth the weighted \text{RobOT} matching to combine accuracy with topology preservation and, in the case of the \text{LDDMM} model, guarantees of invertibility.

\item The spline and the \text{LDDMM} model both result in smooth deformations. But crucially, the \text{LDDMM} model is more suited to large deformations and provides a \textbf{better fit to the corners of the target cube}.
We note that workarounds exist for
the over-smoothing of the second (spline) column:
at the cost of an increased sensitivity to noise,
singular kernels \cite{shepard1968two}
or ridge regression methods \cite{bookstein1989principal}
allow spline models to get a closer fit to the target.
For medical applications, the key benefit of LDDMM
is that it provides strong guarantees on the invertibility
of the deformation:
we refer to Chapter~5 of~\cite{feydy2020geometric} for a
detailed discussion.
In the remainder of this work, we benchmark
both spline and LDDMM deformation models
whenever relevant.
We observe that diffeomorphic LDDMM registrations
perform better on e.g. lung data,
but stress that such comparisons are task-dependent.
\end{itemize}

\subsubsection{Deep feature learning}
\label{sec:suppl_deep_feature_learning}



\textbf{Feature learning on 3D point clouds.}
As discussed above, $(x,y,z)$ coordinates or
standard FPFH features can be good enough to handle simple shapes and deformations.
On challenging settings however, learning task-specific
features is often key to high performance.

In Sec.~\ref{sec:reg_deep_deform} and Suppl.~\ref{sec:suppl_deep_deform}, 
we present our best-performing solution to this problem:
the D-RobOT architecture.
In this section, we discuss an \textbf{alternative approach}
that decouples feature learning and matching.
In practice, 
we did not obtain competitive results with this method.
Nevertheless, using a separate feature extractor
may increase interpretability and 
ease deployment issues in medical scenarios:
we believe that this line of work is worth
pursuing and provide full details on our experiments.


Let us process a synthetic pair of point clouds with $\text{N}$ points in correspondence with each other -- $\text{N}=60,000$ in our experiments for lung registration.
Instead of sampling positive and negative samples, 
which is common for contrastive loss functions,
we choose to rely on $\text{N}\times \text{N}$ correspondence matrices that 
take \textbf{all possible point pairs into account}. 
We use the KeOps library \cite{feydy2020fast} to manipulate 
these objects efficiently, 
with extremely fast run times and without memory overflows.

Specifically, we train our feature extraction network as follows: 
\begin{compactenum}
\item \textbf{Input data.} We assume that we are given two point clouds $(x_1, \dots, x_\text{N})$ and $(y_1, \dots, y_\text{N})$ that are in pairwise correspondence with each other. In our experiments, these are typically the output of a synthetic ``data augmentation'' procedure: the Flying3D objects for the Kitti benchmark and our synthetic lung pairs for the Dirlab benchmark.
\item \textbf{Feature extraction.} We apply the feature extractor (a trainable point neural network) on both point clouds, independently from each other. We retrieve point features $p_i$ and $q_i$ in $\mathbb{R}^\text{D}$ that are respectively associated to the points $x_i$ and $y_i$. 
\item \textbf{Feature normalization.} As discussed in Sec.~\ref{sec:mathematical_background},
we normalize the feature vectors so that $\|p_i\|_{\mathbb{R}^\text{D}} = \|q_i\|_{\mathbb{R}^\text{D}} = 1$. This prevents
the feature extractor from converging to degenerate solutions
and ensures that our hyper-parameters for the RobOT problem of Eq.~(\ref{eq:unbalanced_Sinkhorn}) can be interpreted
as sensible scales in $\mathbb{R}^\text{D}$.
\item \textbf{Source self-similarities.} We compute the $\text{N}\times \text{N}$ correspondence  matrix for the point positions in the source point cloud:
\begin{align}c_{(x_i,x_j)} = \text{softmax}_{j=1}^\text{N}(-\|x_i-x_j\|^2_{\mathbb{R}^3} / 2 \kappa^2) = \frac{\exp(-\|x_i-x_j\|^2_{\mathbb{R}^3} / 2\kappa^2 )}{\sum_{j=1}^\text{N}\exp(-\|x_i-x_j\|^2_{\mathbb{R}^3} / 2\kappa^2)}~,
\end{align}
where the Softmax denotes a mirrored exponential followed by a normalization over the rows of the correspondence matrix while $\kappa > 0$ is a scaling factor. Each row of the matrix $c_{(x_i, x_j)}$ then refers to a probability distribution, a position heatmap with a peak at point $x_i$ whose radius is proportional to $\kappa$.
\item  \textbf{Feature correspondences.} Similarly, we compute a correspondence matrix for the source and target features: 
\begin{align}c_{(p_i,q_j)} = \text{softmax}_{j=1}^\text{N}(-\|p_i-q_j\|^2_{\mathbb{R}^\text{D}})~.
\end{align}
Each row of $c_{(p_i,q_j)}$ is a feature heatmap that indicates how well $p_i$ corresponds to $q_j$.
\item \textbf{Training loss.} Our total loss is the sum over the cross entropies for each row: 
\begin{align}
\text{CE}(c_{(x_i,x_j)}, c_{(p_i,q_j)})=-\sum_{i=1}^\text{N} \sum_{j=1}^\text{N} c_{(x_i,x_j)}\log~c_{(p_i,q_j)}~.
\end{align}
We optimize it by stochastic gradient descent over 
the parameters of the feature extraction network (step 2).
\end{compactenum}

\begin{figure}[!t]
\includegraphics[width=\textwidth]{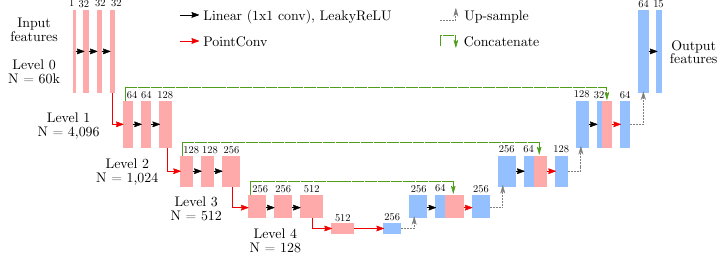}
\caption{
\textbf{Deep feature extractor} for our S-RobOT experiments.
We adapt this feature pyramid network from \text{PointPWC-Net} \cite{wu2019pointpwc}. We implement a four-scale U-net \cite{unet} architecture using: point-wise multi-layer perceptrons on the feature embeddings (with LeakyReLU non-linearities); PointConv~\cite{wu2019pointconv} for downsampling and gathering information from the neighbors; K-Nearest Neighbor interpolation for upsampling.\\
Please note that this architecture relies on
the point coordinates $(x,y,z)$ to define
the PointConv convolution and the upsampling layer.
In our experiments for lung registration,
we use the local vessel radius as our
only input feature: this corresponds
to using a single input channel on the left-most
layer of the figure above.
}
\label{fig:deep_feature_extractor}
\end{figure} 

\textbf{Hyper-parameters.}
For feature learning, we use a U-net structured PointConv~\cite{wu2019pointconv} architecture with 7M parameters
that is illustrated in Fig.~\ref{fig:deep_feature_extractor}.
At the feature learning stage, we set $\kappa$ (described above) to $\sqrt{2}$ mm. For each point we learn a 15 dimensional feature vector of unit length. 
For the \text{RobOT} feature matching, we set the \emph{blur} parameter to 
$\sigma = 0.01$ units in $\mathbb{R}^{15}$ and the \emph{reach} parameter to 
$\rho = +\infty$ (balanced OT). 
As discussed in Suppl.~\ref{sec:suppl_lung}, we benchmark two types of S-RobOT regularization:
\begin{compactenum}
    \item The spline smoothing of Eq.~(\ref{eq:spline}), using a Gaussian (RBF) kernel with a standard deviation of $5$\,mm.
    \item An LDDMM deformation model that we optimize as in Eq.~(\ref{eq:morph_optimization}) using an SGD solver.
    We use a multi-Gaussian-kernel with standard deviations $\{5, 8, 10\}$ mm and weights $\{0.2,0.3,0.5\}$.
\end{compactenum}

\subsection{Additional material on deep deformation prediction}
\label{sec:suppl_deep_deform}

We now provide full details on our D-RobOT architecture
and additional experiments on scene flow estimation
and lung registration.

\subsubsection{Deep registration module}
\label{sec:suppl_prediction_architecture}

{\bf Architecture of the prediction network.} 
In Sec.~\ref{subsec:deep_deformation}, we rely on a modified \text{PointPWC-Net} to act as a predictor $\text{Pred}:(x_i,y_j)\mapsto \theta$. 
This multiscale point neural network takes as input 
the source and target point clouds. 
It returns
a high-dimensional parameter $\theta$
for the deformation model
$\text{Morph}:(\theta, x_i)\mapsto \hat{y}_i$.
In this work, the predicted $\theta$ 
is always a 3D vector field that is supported
by the points $x_i$ (for the raw displacements model)
or by a collection of control points $c_i$ 
that have been generated by farthest point sampling
(for the spline and LDDMM models).

We describe our modifications to the original
PointPWC-Net architecture in Fig.~\ref{fig:deep_reg_module}.
In order to define a network that 
can predict spline and LDDMM parameters: 
\begin{compactenum}
\item We replace the flow prediction layer of \text{PointPWC-Net} by a suitable prediction layer for registration parameters.
\item We use an asymmetric hierarchical architecture that outputs registration parameters for the control points $c_i$. Since the number of control points is often much smaller than the number of points in the input point cloud, this reduces the memory footprint of our network architecture.
\end{compactenum}

\afterpage{
\begin{figure}[!t]
\includegraphics[width=1\textwidth]{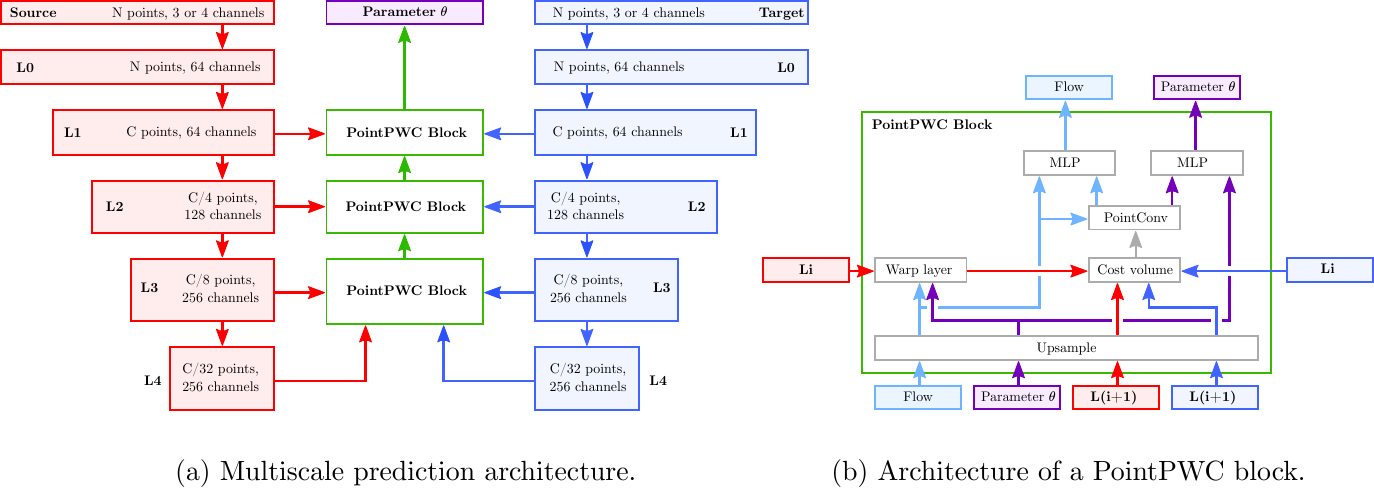}
\caption{\textbf{Architecture of our deep prediction network} $\text{Pred}:(x_i,y_j)\mapsto \theta$, which is adapted from \text{PointPWC-Net} \cite{wu2019pointpwc}. 
The original PointPWC-Net architecture is a multi-scale network
that predicts 3D scene flow (a raw displacement field)
in a coarse-to-fine fashion.
We modify this state-of-the-art architecture
to make it suitable for \textbf{registration
with a task-specific deformation model}
$\text{Morph}:(\theta, x_i)\mapsto \hat{y}_i$. \\
\textbf{a)} First of all, we compute \textbf{feature pyramids}
at scales L0 to L4 using farthest point sampling
and PointConv layers \cite{wu2019pointconv}
on the source and target shapes (left and right columns)
-- see Fig.~5 in \cite{wu2019pointpwc}.
In our experiments, we always use
$\text{C} = 4,096$
or 
$\text{C} = 8,192$
\textbf{control points}.
The key design decision behind the PointPWC-Net architecture
is to run through this pyramid
from the coarsest scale (L4) to the finest one (L0, the original point clouds)
in order to predict the final 3D scene flow: 
we represent this coarse-to-fine prediction
as a stack of \textbf{PointPWC blocks} (central column). \\
\textbf{b)} Those blocks share the same architecture 
(but not the same neural weights):
an \textbf{upsampling} layer that interpolates our 3D vector fields
from the coarser to the finer scale using an inverse distance spline kernel; another \textbf{upsampling} layer that interpolates the pyramid features at the coarsest scale;
a \textbf{warp layer} that deforms the source shape
according to the up-sampled 3D flow, allowing 
the network to focus on \textbf{residual} deformations at each scale;
a \textbf{cost ``volume''} layer,
detailed in Fig.~2 of \cite{wu2019pointpwc},
which relies on K-NN neighborhoods and PointConv layers
to merge the source and target features
into a vector of ``patchwise'' discrepancies
for every point in the (subsampled and warped) source shape;
a \textbf{prediction} PointConv and Multi-Layer Perceptrons (MLP), detailed in Fig.~6 of \cite{wu2019pointpwc}, that turns this vector of discrepancies
into a predicted correction for the up-sampled 3D flow.\\
Please note that this architecture
relies on the point coordinates $(x,y,z)$
in the PointConv and upsampling layers.
For our scene flow experiments,
we use the $(x,y,z)$ coordinates
as input to the network:
this corresponds to ``Source'' and ``Target'' arrays that have 3 input channels.
For our lung registration experiments,
we also add the local vessel radius
as a fourth feature and thus use 4 input channels.
We stress that in our RobOT-based pre-alignment
and post-processing (steps 1 and 3 of the D-RobOT architecture), we rely on the
$(x,y,z)$ coordinates as point features
$p_i$ and $q_j$ in Eq.~(\ref{eq:unbalanced_Sinkhorn}).
When available, the vessel radii
are only used as point weights
$\alpha_i$ and $\beta_j$ in the RobOT problem.\\
In all our experiments, the parameter $\theta$ is a 3D vector field
that has the same memory footprint as a ``raw'' 3D scene flow supported by our control points:
this corresponds to using
3 output channels on the $\text{C}$ control points.
Our main modification to the original PointPWC-Net architecture
is to \textbf{handle this registration parameter} (purple)
\textbf{in parallel with the usual scene flow} (sky blue):
the final prediction module of every PointPWC block
runs \textbf{two estimations in parallel},
which share the same PointConv layer
and are equivalent to Fig.~6 in \cite{wu2019pointpwc}.
}
\label{fig:deep_reg_module}
\end{figure} 
\clearpage
}

{\bf Losses.} We use synthetic data pairs for training: 
for scene flow estimation on \text{Kitti}, we rely on the synthetic \text{Flying3D} dataset; 
for lung registration, we rely on 
the two-scales simulator of Suppl.~\ref{sec:suppl_synth_data}. 
In both cases, we thus have access to ground truth deformations and can rely on a supervised learning strategy. Let us consider a \text{PointPWC-Net} with $\text{L}$ scales, and denote by $(x^l_1, \dots, x^l_{\text{N}_l})$ the subsampled input for the $l$-th scale. 
For training, we compute a multi-scale similarity loss as:
\begin{align}
    \text{Loss}(\hat{y}_i)~=~ \textstyle\sum_{l=0}^{\text{L}-1} W^l \textstyle\sum_i w_i\cdot \|\, \hat{y}^l_i~-~y^l_i\,\|^2_2~,
    \label{eq:pointpwc_multiscale_loss}
\end{align} 
where $l\in \{0,\dots,\text{L}\}$ is a scale 
($l=0$ corresponds to the raw point clouds), 
$W^l$ is a scalar hyper-parameter that we use as a total weight for the $l$-th scale, 
$\hat{y}^l_i$ is the output of the network that corresponds to the flowed $x^l_i$ at the $l$-th scale and $y^l_i$ denotes the ground truth target that corresponds to the same point $x^l_i$ with weight $w_i$. Note that at the finest scale, 
we compute $\hat{y}^0$ using a task-specific deformation model, $\text{Morph}:(\theta,x_i) \rightarrow \hat{y}_i$.

\subsubsection{Experiments on lung vascular trees}
\label{sec:suppl_lung}

\begin{figure}[!p]
\includegraphics[width=1\textwidth]{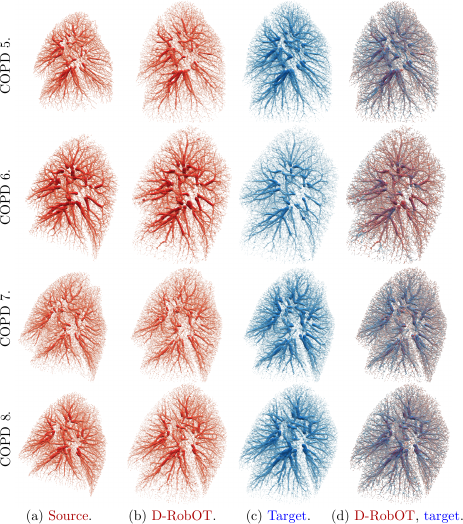}
\caption{\textbf{Additional results for lung vascular trees.} 
Columns refer to:
(a) the source shape (expiration); 
(b) the registration result from \text{D-RobOT} using 
the \text{LDDMM} deformation model;
(c) the target shape (inspiration);
(d) an overlap between
the \text{D-RobOT} registration and the target. 
Each row corresponds to a patient, with names that refer to case IDs 
in the original \text{DirLab} dataset.}
\label{fig:more_example_lung}
\end{figure}

\begin{figure}[!p]
\RawFloats
\centering

\includegraphics[width=.85\textwidth]{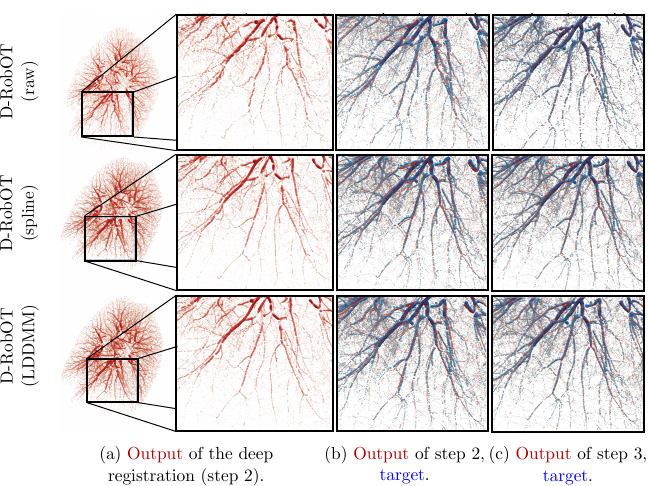}
\caption{\textbf{D-RobOT results for different deformation models 
$\text{Morph}:(\theta, x_i)\mapsto \hat{y}_i$.}
(a) The first two columns correspond to the output of our deep registration module
(without post-processing);
(b) the third column also includes 
the target shape, displayed as a blue point cloud;
(c) the fourth column showcases our final result,
with RobOT-based post-processing.
As detailed in Suppl.~\ref{sec:suppl_lung}, we observe that: (i) the spline and \text{LDDMM} models produce smooth registration results; (ii) before the \text{RobOT} post-processing step, \text{LDDMM} provides a better fit than the spline model; (iii) after post-processing, all three deformation models produce ``sharp'' results, with smoothness guarantees for the spline and LDDMM models.}
\label{fig:lung_compare_model}

\vspace*{.5cm}
\includegraphics[width=1\textwidth]{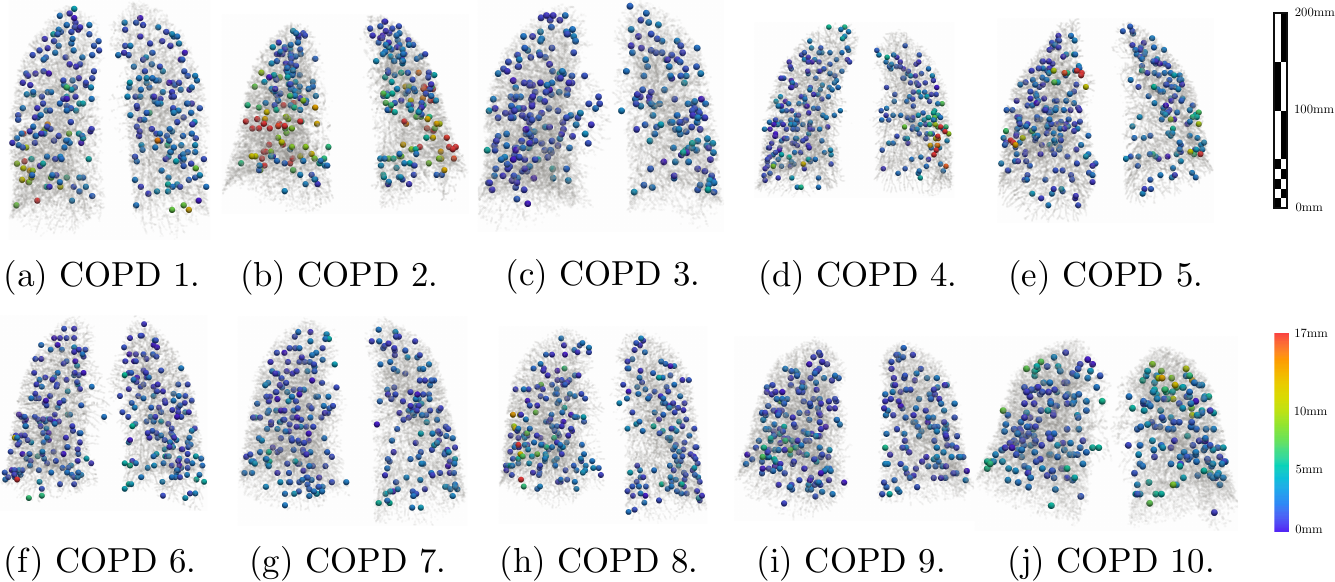}
\caption{\textbf{Distributions of the \text{DirLab}-\text{COPDGene} errors for the 300 expert-annotated corresponding landmarks.} 
Each image refers to a patient, with a title that refers to the 
case ID in the original \text{DirLab} dataset. 
We use color to display the registration error with our best performing model (D-RobOT with LDDMM deformations) on the source landmarks. 
The COPD~2 patient is a clear outlier
in our experiment as in all previous studies 
on the original volumetric data
(\url{https://www.dir-lab.com/Results.html})
-- this may be due to annotation errors.
On the other patients, most of the large errors occur in boundary regions
where acquisition artifacts induce inconsistencies 
between the source and target point clouds:
small vessels are harder to catch in the original 3D volume.}
\label{fig:landmark_error}
\end{figure}

We now provide more details on our experiments for lung registration,
which are illustrated in Fig.~\ref{fig:more_example_lung}
with a visualization of landmark errors in Fig.~\ref{fig:landmark_error}.
Notably, we discuss both supervised and unsupervised training strategies as well as the influence of the deformation model 
$\text{Morph}:(\theta, x_i)\mapsto \hat{y}_i$
on the D-robot architecture.

{\bf Synthetic data vs unsupervised learning.} 
To overcome the lack of dense 3D annotations on real shape data,
we advocate the use of simulated deformations
and synthetic training datasets.
As detailed in Sec.~\ref{subsec:deep_deformation},
we observe that D-RobOT networks are easy to train 
to a high level of accuracy:
RobOT-based post-processing and fine-tuning help
our models to bridge the domain gap
between the synthetic and real distributions of shapes.

We would like to stress that our focus on synthetic
training datasets to the detriment of e.g. unsupervised approaches
results from \textbf{careful and extensive experiments}.
In the lead-up to the publication of this work,
we tried several competing approaches to train
our registration networks:
supervised learning with dense correspondences
on synthetic data;
``unsupervised'' learning on unannotated 
point clouds using geometric loss functions between point sets;
a mix of both approaches.
In practice, supervised learning on synthetic
data clearly emerged as the most practical option
for challenging registration tasks.
Let us briefly explain why.

\textbf{Unsupervised loss functions.}
When dense pointwise correspondences are not available,
a popular strategy to train registration methods
is to rely on permutation-invariant
loss functions between point clouds.
We refer to Chapters 3 and 4 of \cite{feydy2020geometric}
for an introduction to the topic.

Since our PVT1010 dataset contains 1,000 pairs of
lung vessel trees that are in correspondence
with each other (inspiration/expiration)
but for which no expert-annotated landmarks are available,
we are in a perfect situation to try out these tools.
We thus attempted to train our networks
using local Laplacian matching~\cite{wu2019pointpwc}, 
Maximum Mean Discrepancies, 
Gaussian Mixture Models (\text{GMM}) 
as well as Wasserstein distances \cite{feydy2020geometric}
on $(x, y, z)$ coordinates.
Unfortunately, we 
\textbf{never succeeded in converging to a competitive 
level of accuracy}. 
We believe that this is due to the 
\textbf{complex geometric structure 
of the lung vascular trees}, which
are significantly more intricate than the
clean point clouds and surface meshes on which
these methods are usually tested \cite{feydy2017optimal,feydy2018global}.

\textbf{Combining synthetic (supervised) and real (unsupervised) data.}
Going further,
we also tried to use a \textbf{mixed strategy}:
in our training dataset,
we combined synthetic deformations of real source shapes (that can be handled using a mean square error) 
with genuine target point clouds (that can be handled using e.g. the Wasserstein distance). 
We expected that this combination would narrow the \textbf{domain gap} between our synthetic deformations and the real breathing movement. In practice, this strategy produced \textbf{indecisive results}:
\begin{itemize}
    \item On the one hand, assuming that we only have access to a \textbf{simple  deformation model} (e.g. a single-scale random field), this mixed training strategy improves the accuracy of our registrations. This is especially true when the synthetic deformations are not diverse enough. 
    \item On the other hand, assuming that we have access to the more \textbf{realistic and expressive} deformation model of Suppl.~\ref{sec:suppl_synth_data}, the introduction of real but not annotated pairs in the training loop proves \textbf{slightly detrimental} to performance. 
    We observe a small increase of about 0.2mm 
    in landmark Root Mean Squared Error (RMSE).
\end{itemize}
For the sake of \textbf{simplicity}, we thus trained our final \text{D-RobOT} network entirely on synthetic deformations.
We found that using \text{RobOT} as a \textbf{final post-processing} layer in the \text{D-RobOT} pipeline is enough to address prediction errors effectively and compensate for a small domain gap between synthetic and real deformations.

{\bf Raw displacements vs spline and \text{LDDMM} regularizations.}
In Fig.~\ref{fig:lung_compare_model}, we further compare the \text{D-RobOT} results for three different deformation models.
Our main observations are that:
\begin{itemize}
    \item A deep prediction module that returns raw displacements $\Delta x_i$
    with a naive deformation model $\text{Morph}:(\theta=\Delta x_i, x_i)\mapsto \hat{y}_i = x_i+\Delta x_i$
    may produce non-smooth registration results.
    This holds even when the network is trained entirely 
    on smooth, synthetic deformations.
    On the other hand, the regularizing 
    spline and \text{LDDMM} models always
    result in smooth deformations.
    This vindicates the use of \textbf{explicit regularizing models}
    in safety-critical applications:
    we cannot trust our networks to ``learn smoothness properties''
    from the data.
    \item According to Fig.~\ref{fig:lung_compare_model} and Tab.~\ref{tab:lung_compare_post}, the \text{LDDMM} model captures large deformations better than the spline model. However, we obtain similar performance with both spline and \text{LDDMM} models after the (affordable) \text{RobOT} post-processing step: \textbf{fine-tuning with \text{RobOT} alleviates the need for complex and expensive deformation models}.
\end{itemize}

{\bf Experiments.}
We now provide detailed hyperparameters for the \text{D-RobOT} architecture in our lung registration experiments. As detailed in Sec.~\ref{sec:reg_deep_deform}, the full model applies successively: an affine S-RobOT pre-alignment; a deep non-parametric registration; and a spline S-RobOT post-processing.

\textbf{1. Affine registration}. 
For affine pre-alignment,
we use the \text{S-RobOT} model of Eq.~\eqref{eq:affine} with raw $(x,y,z)$ coordinates as input features in $\mathbb{R}^3$. The vessel radii are taken into
account as point weights $\alpha_i$ and $\beta_j$.
In the RobOT problem of Eq.~(\ref{eq:unbalanced_Sinkhorn}),
we set the \emph{blur} parameter to $\sigma=\text{1\,mm}$ and the \emph{reach} parameter to $\tau=+\infty$ (balanced OT with strong constraints on the marginals).

\textbf{2. Deep non-parametric registration}. We evaluate three deformation models $\text{Morph}:(\theta, x_i)\mapsto \hat{y}_i$: 
\begin{compactenum}
\item \textbf{Raw displacements}
correspond to the case where the parameter
$\theta = \Delta x_i$
is a 3D vector field that is supported
by the point $x_i$ and the deformation
model is the addition:
\begin{align}
    \text{Morph} ~:~
    (\Delta x_i, x_i) \in
    \mathbb{R}^{\text{N}\times 3}
    \times 
    \mathbb{R}^{\text{N}\times 3}
    ~\mapsto~ 
    x_i + \Delta x_i 
    \in
    \mathbb{R}^{\text{N}\times 3}~.
\end{align}
As a regularization penalty,
we use the squared Euclidean norm:
\begin{align}
\text{Reg}(\Delta x_i) = 
\tfrac{1}{\text{N}}
\textstyle\sum_{i=1}^{\text{N}} \|\Delta x_i\|^2_{\mathbb{R}^3}~.
\end{align}

\item The \textbf{spline} model corresponds
to the case where the parameter
$\theta = \Delta c_l$
is a 3D vector field that is supported
by a set of control points 
$(c_1, \dots, c_\text{C})$ in $\mathbb{R}^3$.
As a deformation model, we use a simple kernel smoothing:
\begin{align}
    \text{Morph} ~:~
    (\Delta c_l, x_i) \in
    \mathbb{R}^{\text{C}\times 3}
    \times 
    \mathbb{R}^{\text{N}\times 3}
    ~\mapsto~ 
    x_i + 
    \textstyle\sum_{l=1}^{\text{C}} k(x_i, c_l) \Delta c_l
    \in
    \mathbb{R}^{\text{N}\times 3}~.
\end{align}
As a regularization penalty,
we use the squared Euclidean norm: 
\begin{align}
\text{Reg}(\Delta c_l) = 
\tfrac{1}{\text{C}}
\textstyle\sum_{l=1}^{\text{C}} \|\Delta c_l\|^2_{\mathbb{R}^3}~.
\end{align}

For our lung experiments, we always use
a multi-Gaussian kernel with standard deviations $\{3,6,9\}$ mm and weights $\{0.2,0.3,0.5\}$.
With the coordinates of points 
$x$ and $y$ in millimeters, this corresponds to
the positive definite kernel function:
\begin{align}
    k(x,y) ~&=~ 
    0.2 \cdot \exp\big[ -\|x-y\|^2_{\mathbb{R}^3} / (2 \cdot 3^2)\big]
    ~+~
    0.3 \cdot \exp\big[-\|x-y\|^2_{\mathbb{R}^3} / (2 \cdot 6^2)\big] \\
    ~&+~
    0.5 \cdot \exp\big[-\|x-y\|^2_{\mathbb{R}^3} / (2 \cdot 9^2)\big]~. \nonumber
\end{align}

\item The \textbf{diffeomorphic LDDMM} model is equivalent to a 
spline deformation with continuous time.
Notably, we use the same kernel $k(x,y)$ as in the spline model above.
As detailed in Suppl.~\ref{sec:suppl_lddmm} and Chapter~5 of \cite{feydy2020geometric},
the parameter $\theta = m^0_l$
is a 3D vector field that is supported
by a set of control points 
$(c_1, \dots, c_\text{C})$ in $\mathbb{R}^3$.
As a deformation model, we use an iterative kernel smoothing
and deformation layer
that corresponds to a numerical integration scheme
for the Ordinary Differential Equation (\ref{eq:lddmm_hamiltonian}).
More specifically, we use the differentiable Dopri-5 layer
that is provided by the TorchDiffEq library \cite{chen2018neuralode,chen2021eventfn},
which implements the Dormand-Prince or Runge-Kutta 4(5) integrator \cite{dormand1980family}
with 20 time steps from time $t=0$
to time $t=1$.
For a simpler implementation,
we refer to the Euler integration scheme
that is detailed in~\cite{feydy2020geometric}, Algorithm 5.6.
As a regularization term, we use the squared kernel norm:
\begin{align}
\text{Reg}(\Delta c_l) = 
\tfrac{1}{\text{C}}
\textstyle\sum_{l=1}^{\text{C}} 
\textstyle\sum_{s=1}^{\text{C}}
k(c_l, c_s)
\langle m^0_l, m^0_s\rangle_{\mathbb{R}^3}~.
\end{align}

\end{compactenum}

For the \textbf{deep prediction network} $\text{Pred}:(x_i,y_j)\mapsto \theta$, we use the 4-scale hierarchical architecture that is described in Suppl.~\ref{sec:suppl_prediction_architecture} and set 
the scale weights $W^l$ to $\{1.0,0.8, 0.4, 0.2\}$. 
We compute the 4-scale decompositions of the point clouds using
Farthest Point Sampling.
For all models, 
we promote a close fit to the target shape and weight the regularization term $\text{Reg}(\theta)$ by 1/100 in the training loss: 
\begin{align}
    \text{Loss}(x_i, y_i, \theta)
    ~=~
    \tfrac{1}{100} \text{Reg}(\theta)
    ~+~
    \underbrace{
    \textstyle\sum_{l=0}^{\text{L}-1} W^l \textstyle\sum_i w_i\cdot \|\, \hat{y}^l_i(\theta, x_i)~-~y^l_i\,\|^2_{\mathbb{R}^3}}_{\text{Multiscale loss of Eq.~(\ref{eq:pointpwc_multiscale_loss})}}~.
    \label{eq:training_loss}
\end{align}
We recall that in the expression above:
\begin{compactenum}
\item The weights $w_i$ are attached to the points $x_i$
      and are proportional to the vessel radius.
\item The weights $W^0$, $W^1$, $W^2$ and $W^3$ put a strong emphasis on the finest scales.
\item Each point $y_i^l$ corresponds to the ground truth target for $x_i$ at scale $l$.
\item Each point $\hat{y}^l_i(\theta, x_i)$ corresponds to the output
of our prediction and deformation networks at scale $l$ for $x_i$.
As detailed in Fig.~\ref{fig:deep_reg_module},
the finest-scale output
$\hat{y}^0_i(\theta, x_i) = \text{Morph}(\theta, x_i)$
corresponds to our model-based deformation;
the larger-scale predictions
$\hat{y}^1_i(\theta, x_i)$,
$\hat{y}^2_i(\theta, x_i)$ and
$\hat{y}^3_i(\theta, x_i)$ correspond to raw displacements
and are only used in Eq.~(\ref{eq:training_loss})
to make the training easier \cite{wu2019pointpwc}.
\end{compactenum}

\textbf{3. Postprocessing}. To fine-tune our registration, we use the spline \text{S-RobOT} deformation of Eq.~(\ref{eq:spline}) with $(x,y,z)$ coordinates as input features in $\mathbb{R}^3$. We set the \emph{blur} parameter to $\sigma = \text{0.1 mm}$ and the \emph{reach} parameter to $\tau = \text{10 mm}$.
For smoothing,
we use the vessel-preserving anisotropic kernel $k(x,y)$ of Sec.~\ref{sec:suppl_synth_data}
with a kernel scale $s_{\text{local}} = \text{8 mm}$. 
In order to get a closer fit to the target, 
we apply this post-processing step twice for all lung registration results.

\begin{table}[p]
\RawFloats
\centering

\scalebox{.95}{
\begin{tabular}{|lccc|}
  \hline
 & \textbf{D-RobOT (raw)} & \textbf{D-RobOT (spline)} & \textbf{D-RobOT (LDDMM)} \Tstrut \\
\textbf{Post-processing} & mm $\downarrow$ & mm $\downarrow$ & mm $\downarrow$ \Bstrut \\ \hline
Without post-processing & 
3.46 (3.13) & 4.45 (4.10) & 3.33 (3.09)\Tstrut\\ 
NN projection & 
3.33 (2.96) & 3.32 (3.05) & 3.31 (2.96)\\ 
RobOT & 
3.35 (3.04) & 3.18 (3.01) & 3.19 (3.01)\\ 
NN projection + Smoothing &
\textbf{3.32 (2.95)} &3.08 (2.82) &2.95 (2.64) \\ 
RobOT + Smoothing &
3.33 (2.99) & \textbf{2.94 (2.71)}  & \textbf{2.83 (2.40)}\Bstrut\\
\hline 
\end{tabular}}

\vspace*{.25cm}
\caption{\textbf{Ablation study on the post-processing module.} 
We assess the influence of the fine-tuning step in the D-RobOT model.
As an addendum to Tab.~\ref{fig:dirlab},
we display the Root Mean Squared Error in millimeters
(median of the 10 subjects in parentheses)
for expert-annotated landmarks on the 10 Dirlab lung pairs,
that we use as a test set for PVT1010. 
The first row corresponds to the output of the deep registration
module, without any fine-tuning.
The second row corresponds to a nearest neighbor projection
on the target point cloud.
The third row corresponds to the 
``raw'' RobOT matching of Eq.~\ref{eq:robot_matching},
extrapolated to the landmarks using a Nadaraya-Watson
spline with an isotropic Gaussian kernel of deviation
$\sigma = 0.5\,\text{mm}$.
In the fourth and fifth rows, we use a vessel-preserving
anisotropic kernel to smooth a nearest neighbor
projection (fourth row)
and our RobOT matching (fifth row).}
\label{tab:lung_compare_post}

\vspace*{.5cm}
\renewcommand{\tabcolsep}{3pt}
  \begin{center}
  \scalebox{.9}{
  \begin{tabular}{|lcccccc|}
    \hline
    ~~ & ~\textbf{Pre-}~ & ~\textbf{Deep}~ & ~\textbf{Post-}~&~\textbf{Number of}~&~\textbf{Number of}~ & ~\textbf{RMSE} ~ \Tstrut\\
    ~\textbf{Method}~ & ~\textbf{alignment}~ & ~\textbf{prediction}~ & ~\textbf{processing}~&~\textbf{points}~&~\textbf{control points}~ & ~mm $\downarrow$ ~
    \Bstrut \\
    \hline
    Input data & & None & & 60k & ----   & 23.32\Tstrut\\
    Affine & \cmark & None & & 60k & ----   & 10.31\Bstrut\\
    \hline
    \multirow{3}{*}{FlowNet3D} 
 &  & FLOW & & 30k & 4,096   & 8.20\Tstrut\\
     &  \cmark& FLOW & & 30k &4,096 &  7.08\\
     
     &  \cmark& FLOW &\cmark & 30k &4,096 & 6.51\Bstrut\\
    \hline
     
    \multirow{5}{*}{D-RobOT}  &  \cmark& \text{PWC}$^*$& & 30k & 4,096 & 4.10\Tstrut\\
     
      &  & \text{PWC$^*$}& & 60k & 4,096 & 4.64 \\
      &  \cmark& \text{PWC$^*$}& & 60k & 4,096 & 3.82 \\
    
      & \cmark& \text{PWC$^*$}& & 60k & 8,192 & 3.33  \\
      & \cmark& \text{PWC$^*$}& \cmark& 60k & 8,192 & \textbf{2.83}\Bstrut\\
    \hline
  \end{tabular}}
  \end{center}
  \caption{\textbf{Ablation study for the \text{D-RobOT} model (with LDDMM deformations) on the lung registration task.} We benchmark the Root Mean Squared Error
  on the 10 $\times$ 300 pairs of DirLab landmarks for a wide range of configurations.
  We investigate the influence of affine \text{S-RobOT} pre-alignment (column 2);
  spline S-RobOT fine-tuning (column 4); 
  the sampling rate for the input data (column 5);
  and the number of control points for the LDDMM deformation model (column 6).
  As a backbone architecture for the deep prediction network,
  we use a \text{FlowNet3D} (rows 3-5) and the modified
  \text{PointPWC-Net} of Fig.~\ref{fig:deep_reg_module} (rows 6-10).}
  \label{tab:ablation}

\vspace*{.5cm}

\scalebox{.9}{
\begin{tabular}{|p{0.3cm}p{2.5cm}cccccccc|}
  \hline 
& \textbf{Method} &
\textbf{Prealign} &
\textbf{Post} &
\textbf{EPE3D} & \textbf{Acc3DS} & \textbf{Acc3DR} & \textbf{Outliers3D} & \textbf{EPE2D} & \textbf{Acc2D} \\
& 
&
&
& cm $\downarrow$ 
& \% $\uparrow$ 
& \% $\uparrow$ 
& \% $\downarrow$ 
& px $\downarrow$ 
& \% $\uparrow$\\\hline 
\parbox[t]{1mm}{\multirow{5}{*}{\rotatebox[origin=c]{90}{\textbf{8192 points}}}} 
& \text{PWC}~\cite{wu2019pointpwc}  &  & &
6.78 & 78.58 & 90.30 & 23.61 & 2.6484 & 82.36\Tstrut\\
& \text{PWC}  & \cmark  & & 
4.62 & \best{83.24} &94.91 & \best{19.61} &2.1411 &86.82 \\
& \text{PWC}  &\cmark  &\cmark & 
4.76 & 82.16  &  94.53  & 20.10  & 2.0716  & \best{87.59}\\
& \text{D-RobOT}  &\cmark  & & 
4.63 & 79.13 & \best{95.30} & 21.43 & 2.3930 & 82.93\\ 
& \text{D-RobOT} & \cmark &\cmark & 
\best{4.55} &  80.27 &  94.85 &  20.65 &  2.1404 &  85.24\\
\hline 

\parbox[t]{1mm}{\multirow{5}{*}{\rotatebox[origin=c]{90}{\textbf{30k points}}}} 
& \text{PWC}~\cite{wu2019pointpwc}  &  & &
7.90 &61.33  &89.75  &29.36  &4.0558  &  65.35\\
 & \text{PWC}  & \cmark & & 
 5.94   & 72.62   & 92.26    &25.79    &3.1817   & 72.82 \\
& \text{PWC} & \cmark &\cmark & 
3.99  & 86.44  & 95.02  & 18.59 & 1.9592 & 86.09\\
& \text{D-RobOT}  &\cmark  & &
2.94 & 92.86 & 98.59 & 16.05 & 1.5733 & 91.86\\
& \text{D-RobOT}  & \cmark & \cmark & 
\bestdense{2.15}   & \bestdense{95.74}   & \bestdense{98.96}   & \bestdense{12.93}   & \bestdense{1.1118}   & \bestdense{95.66}\\
\hline
\end{tabular}}
\vspace*{.25cm}
\caption{\textbf{Ablation study on the modules of \text{D-RobOT} (with spline deformations)} and the influence of the point sampling rate,
performed on the \textbf{57 largest scenes from the \text{Kitti} dataset}. As detailed in Suppl.~\ref{sec:suppl_kitti},
these results correspond to average values on the 57 (out of 142)
pairs of 3D scenes that are sampled with more than 30k points
per frame in the original Kitti dataset.
This allows us to focus our evaluation on the most
challenging 3D scenes, with under-sampling artifacts:
as expected, performance is lower than in the
``full'' Kitti benchmark of Fig.~\ref{fig:kitti}.}
\label{tab:non_oversample}

\end{table} 

\clearpage

\subsubsection{Experiments on \text{Kitti}}
\label{sec:suppl_kitti}

We now provide details for our experiments
on scene flow estimation, which are illustrated 
in Fig.~\ref{fig:kitti_visual}.
Notably, we discuss the influence of the sampling rate for 
the source and target point clouds
and provide an extensive ablation study for the modules
of the D-RobOT model. 

{\bf Influence of the sampling rate.} We follow the pre-processing strategy of  \text{PointPWC-Net}~\cite{wu2019pointpwc}:
\begin{compactenum}
\item We train on points whose depth with respect to the acquisition device is smaller than 35 m.
\item We sample points from both of the source and target frames at random, in a non-corresponding manner.
\end{compactenum} 
In the main manuscript, we report results when the source and target point clouds
are sampled with either 8,192 or 30k points at a time. 
We note that in a similar situation, when sampling 32,768 points from each frame, \text{HPLFlowNet}~\cite{gu2019hplflownet} reported an average \text{EPE3D} value of 10.87 cm over all 142 pairs of the Kitti dataset. In comparison, our D-RobOT model reaches a much higher accuracy (\text{EPE3D} $=$ 2.23 cm) with only 30,000 points per frame. 

\afterpage{
\begin{figure}[!p]
\RawFloats
\includegraphics[width=1\textwidth]{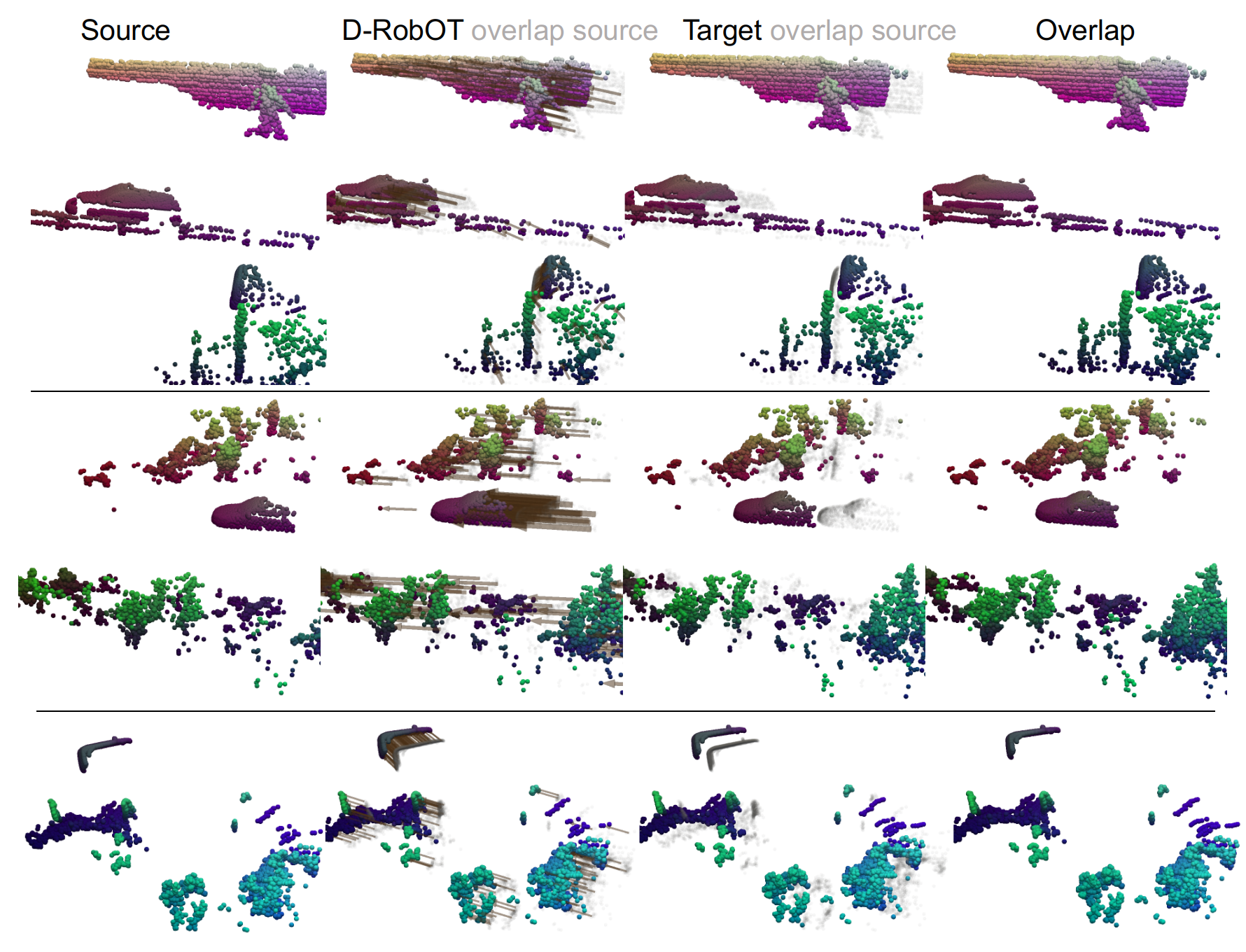}
\caption{\textbf{Additional results for scene flow estimation on \text{Kitti}.}
From left to right, the columns refer to: the source shape; the registration result for \text{D-RobOT} with a spline deformation model (the source point cloud is displayed in gray in the background, flows are displayed as brown arrows); the target shape (the source point cloud is displayed in gray in the background); and the \text{D-RobOT} result overlapped with the target. Each row corresponds to a single pair of 3D frames.}
\label{fig:kitti_visual}

\vspace{.5cm}
\includegraphics[width=1\textwidth]{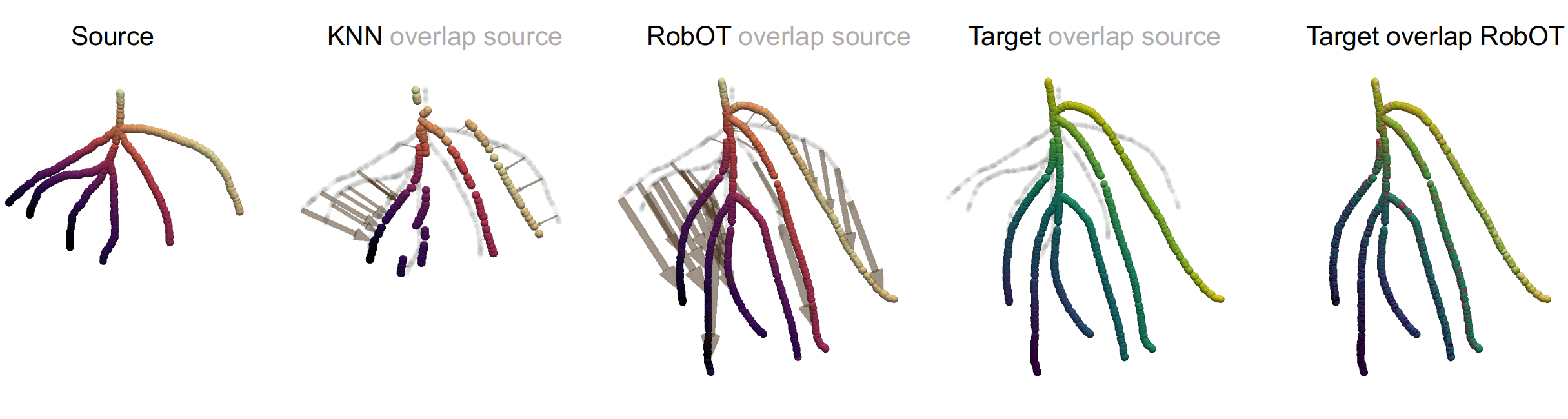}
\caption{\textbf{RobOT matching vs Nearest Neighbor projection on a synthetic local vessel tree.}
From left to right, the columns refer to: the source shape; the registration result for NN projection and \text{RobOT} followed by a local smoother (the source point cloud is displayed in gray in the background, flows are displayed as gray arrows); the target shape (the source point cloud is displayed in gray in the background); and the \text{D-RobOT} result overlapped with the target.}
\label{fig:nn_vs_robot}
\end{figure} 
\clearpage
}

\textbf{Working with the 57 largest scenes from Kitti.}
We note that the number of points per 3D frame varies 
widely in the \text{Kitti} dataset. 
To investigate the potential negative 
effects from under-sampling, we evaluate our models separately 
on pairs where both frames contain more than 30k points: 
57 out of the 142 Kitti scene pairs meet this requirement. 
When dealing with those 57 ``most populated'' frames, 
sub-sampling the scene to 30k points 
induces a net loss of geometric information.

We display our results for this subset of Kitti in Tab.~\ref{tab:non_oversample}: unsurprisingly, performance is lower than in Fig.~\ref{fig:kitti},
which reports results on the full Kitti dataset and thus includes
the 85 pairs of frames for which 30k points are enough to work at full resolution.
Nevertheless, our conclusions remain consistent: 
the D-RobOT method outperforms \text{PointPWC-Net}~\cite{wu2019pointpwc} 
for both sampling rates. 

\textbf{Ablation study.}
Going further, we analyze the contributions of the pre-alignment and
post-processing modules in the D-RobOT model. 
Our results are summarized in Tab.~\ref{tab:non_oversample}
and can be described as follows:
\begin{itemize}
    \item The \textbf{pre-alignment module} improves results for 
    \textbf{both sampling rates}.
    Since rigid transformations have few degrees of freedom,
    the estimation of Eq.~(\ref{eq:rigid}) does not rely on
    fine details.
    \item In sharp contrast, the \textbf{post-processing module} only improves 
    results for \textbf{high-resolution points}.
    In practice, our \text{RobOT}-based post-processing behaves as a fast local fitting to the target point cloud. 
    In the upper half of the table, the target point cloud and the set of 
    ``ground truth'' final positions for the source points have 
    small overlap due to the small number of sampling points for these complex scenes: a local ``pixel-perfect'' fine-tuning is detrimental to performance.
    In the lower half of the table, 
    we increase the sampling rate to 30k points per frame:
    the target point cloud and the set of ground-truth final 
    positions for the source points have a much higher overlap.
    This allows the \text{RobOT} post-processing to increase
    the accuracy of the full model.
\end{itemize}

{\bf Experiments.}
We now provide detailed hyperparameters for the \text{D-RobOT} architecture in our Kitti experiments. As detailed in Sec.~\ref{sec:reg_deep_deform}, the full model applies successively: a rigid S-RobOT pre-alignment; a deep non-parametric registration; and a spline S-RobOT post-processing.

\textbf{1. Rigid registration}. 
For rigid pre-alignment,
we use the \text{S-RobOT} model of Eq.~\eqref{eq:rigid} with raw $(x,y,z)$ coordinates as input features in $\mathbb{R}^3$. The vessel radii are taken into
account as point weights $\alpha_i$ and $\beta_j$.
In the RobOT problem of Eq.~(\ref{eq:unbalanced_Sinkhorn}),
we set the \emph{blur} parameter to $\sigma=\text{1\,m}$ and the \emph{reach} parameter to $\tau=+\infty$ (balanced OT with strong constraints on the marginals).

\textbf{2. Deep non-parametric registration.} 
We use the prediction architecture and deformation
models of Suppl.~\ref{sec:suppl_lung}.
Our spline deformation model is based on a multi-Gaussian-kernel with standard deviations $\{20,40,60\}$\,cm as and weights $\{0.2,0.3,0.5\}$.
With the coordinates of points the
$x$ and $y$ in centimeters, this corresponds to
the positive definite kernel function:
\begin{align}
    k(x,y) ~&=~ 
    0.2 \cdot \exp\big[ -\|x-y\|^2_{\mathbb{R}^3} / (2 \cdot 20^2)\big]
    ~+~
    0.3 \cdot \exp\big[-\|x-y\|^2_{\mathbb{R}^3} / (2 \cdot 40^2)\big] \\
    ~&+~
    0.5 \cdot \exp\big[-\|x-y\|^2_{\mathbb{R}^3} / (2 \cdot 60^2)\big]~. \nonumber
\end{align}

\textbf{3. Postprocessing}. To fine-tune our registration, we use the spline \text{S-RobOT} deformation of Eq.~(\ref{eq:spline}) with $(x,y,z)$ coordinates as input features in $\mathbb{R}^3$. We set the \emph{blur} parameter to $\sigma = \text{5 cm}$ and the \emph{reach} parameter to $\tau = \text{80 cm}$.
For smoothing,
we use the vessel-preserving anisotropic kernel $k(x,y)$ of Sec.~\ref{sec:suppl_synth_data}
with a kernel scale $s_{\text{local}} = \text{80 cm}$. 


\subsection{RobOT matching vs nearest neighbor projection.} 
\label{sec:robot_vs_nn}

As detailed in Sec.~\ref{sec:robot_matching},
we use the RobOT matching of Eq.~(\ref{eq:robot_matching})
as a plug-in replacement for nearest neighbor (NN) projection.
At an affordable computational cost,
our RobOT layer enforces a local conservation of the point density:
this geometric prior is relevant for many
registration tasks and mitigates
accumulation artifacts that are commonly found 
in nearest neighbor projections
\cite{feydy2018global,feydy2020geometric}.
In accordance with the theory, 
we thus observe that RobOT matching
outperforms nearest neighbor projection in
a wide range of settings.

\textbf{Local translations.}
When the point features $p_i$ and $q_j$
correspond to the $(x,y,z)$ coordinates in $\mathbb{R}^3$,
a key property of the Monge-Brenier map
that is defined by Eq.~(\ref{eq:robot_matching})
is that it perfectly retrieves translations
and dilations
\cite{brenier1991polar,peyre2019computational,feydy2020geometric}.
Scene flow estimation is often close
to this best case scenario for RobOT theory:
after we remove points that correspond to the ground
(a common pre-processing for this task),
most of the displacements can be explained 
as local translations and small rotations 
of solid objects (e.g. cars or trees)
in the frame of reference of the acquisition device.

On the Kitti benchmark of Fig.~\ref{fig:kitti}, 
we observe that a simple 
RobOT matching already performs very well
for 3D scene flow estimation.
Remarkably, and without any training,
\textbf{vanilla RobOT matching} on high-resolution point clouds 
(30k points per frame) \textbf{outperforms most 
pre-existing deep learning methods} on medium-resolution
point clouds (8,192 points per frame)
in terms of speed, memory usage and accuracy.
This is strong evidence that geometric methods
deserve more attention from the computer vision community.

\textbf{Complex structures.}
Going further, the mass distribution constraint of Eq.~(\ref{eq:unbalanced_Sinkhorn})
is useful to register \textbf{complex shapes with appendices and branches}.
This property has been discussed in depth in previous works:
we refer to e.g. \cite{feydy2017optimal} for the matching of
five-fingered hand surfaces. The (soft) constraints
on the marginals of the transport plan
promote a bijective matching of the fingers,
preventing the thumb and the index of the source
shape from being both projected on the thumb of the target.

In Fig.~\ref{fig:nn_vs_robot}, we provide a similar example
for the local registration of branching vessels.
We create a synthetic pair of source and target vessels with 
$\text{N}=1,000$ and $\text{M}=1,200$ 
points respectively. We normalize their $(x,y,z)$ coordinates to be contained in 
the unit cube $[0,1]^3$ and assign uniform weights 
$\alpha_i=1$ and $\beta_j=1$ to each point
-- we work in a realistic unbalanced scenario.
We compute the raw NN matching using a simple
nearest neighbor projection from the source onto the target in $\mathbb{R}^3$.
For the RobOT matching, we also use raw $(x,y,z)$ coordinates
and the squared Euclidean metric;
we set the \emph{blur} parameter to $0.0005$ units 
in $\mathbb{R}^3$ and the \emph{reach} parameter to $1$ unit (unbalanced OT).
As detailed in Suppl.~\ref{sec:suppl_synth_data},
we regularize both of the NN and RobOT matchings using an anisotropic
muti-Gaussian-kernel with standard deviations $\{0.03,0.05, 0.07\}$ 
and weights $\{0.2,0.3, 0.5\}$.

\textbf{As a post-processing step.}
As discussed above, RobOT matching
is good at handling local translations, dilations
and small free-form deformations.
We propose to use it as a fast post-processing
step in our D-RobOT architecture, presented in Sec.~\ref{sec:reg_deep_deform}.
In order to validate its utility,
we perform an ablation study on the PVT1010/DirLab dataset
for lung registration.

We benchmark increasingly suitable fine-tuning layers:
no fine-tuning;
nearest neighbor projection;
RobOT matching;
nearest neighbor projection with vessel-preserving smoothing;
RobOT matching with vessel-preserving smoothing.
Our results are detailed in Tab.~\ref{tab:lung_compare_post}:
when used in combination with a smooth deformation model,
these layers result in an increasingly higher accuracy.
This confirms the main findings of our work:
including \textbf{sensible geometric priors}
in a modular deep learning architecture
is the key to reliable state-of-the-art performance.

\subsection{Computational Resources}
\label{sec:comp_reas}

For the evaluation of time and memory cost in Fig.~\ref{fig:kitti}, we compare all models on a Ubuntu server with a single 24GB NVIDIA Quadro RTX 6000 graphics processing unit (GPU) and a 10-core Intel(R) Xeon(R) Silver 4114 CPU @ 2.20GHz. 
We trained all of our networks on a single 24GB NVIDIA RTX 3090 GPU, except for the training of \text{PointPWC-Net} with 30k sampled points that was performed on a 48GB NVIDIA RTX A6000 GPU. 

\subsection{Potential for negative societal impact} 
\label{sec:supply_potential_for_negative}

Point cloud registration is a fundamental low-level task in computer vision and computer graphics. As a consequence, negative societal impact of our work may arise in a wide range of use cases. A first example is that of point cloud registration for autonomous driving:
algorithm failure could lead to incorrect driving decisions. Similar concerns may arise when using these approaches to register facial scans for the purpose of person identification. In general, while our manuscript demonstrates improved performance and robustness over competing approaches, the possibility for registration failures should always be considered when applying our models in safety critical environments.

\end{document}